\documentclass[12pt]{article}
\usepackage{cite}
\usepackage{amsmath,amssymb,amsfonts}
\usepackage{algorithmic}
\usepackage{graphicx}
\usepackage{textcomp}
\usepackage{color}
\usepackage{booktabs}
\usepackage{algorithm}  
\usepackage{algorithmic}
\usepackage{booktabs}
\usepackage{csvsimple}
\usepackage{float}
\usepackage{longtable}
\usepackage{rotating}
\usepackage{lscape}
\usepackage{subcaption}
\usepackage{abstract}
\providecommand{\keywords}[1]{\textbf{\textit{keywords---}} #1}
\def\BibTeX{{\rm B\kern-.05em{\sc i\kern-.025em b}\kern-.08em
    T\kern-.1667em\lower.7ex\hbox{E}\kern-.125emX}}
\usepackage{hyperref}

\usepackage{scicite}

\usepackage{times}

\newcounter{lastnote}

\title{Foldover Features for Dynamic Object Behavior Description in Microscopic Videos}

\author
{Xialin Li $^{1}$, Chen Li $^{1\ast}$ and Wenwei Zhao $^{1}$
\\
\normalsize{$^{1}$Microscopic Image and Medical Image Analysis Group,}\\
\normalsize{MBIE College, Northeastern University, 110169,}\\
\normalsize{Shenyang, PR China}\\
\normalsize{$^\ast$Corresponding author: Chen Li  (e-mail: lichen201096@hotmail.com).}
}

\date{}

\begin{document} 


\baselineskip24pt


\maketitle


\begin{abstract}
  Behavior description is conducive to the analysis of tiny objects, similar objects, objects with weak visual information and objects with similar visual information, playing a fundamental role in the identification and classification of dynamic objects in microscopic videos. To this end, we propose foldover features to describe the behavior of dynamic objects. First, we generate foldover for each object in microscopic videos in X, Y and Z directions, respectively. Then, we extract foldover features from the X, Y and Z directions with statistical methods, respectively. Finally, we use four different classifiers to test the effectiveness of the proposed foldover features. In the experiment, we use a sperm microscopic video dataset to evaluate the proposed foldover features, including three types of 1374 sperms, and obtain the highest classification accuracy of 96.5\%.
\end{abstract}


\section{Introduction}
\label{sec:introduction}
In the field of computer vision, because a video is made up of many frames~\cite{Schultz-1996-EOH} and video analysis is basically image analysis. In addition, we tend to focus on a specific target object or class of video rather than the whole video. Hence, image feature extraction is very important for video analysis~\cite{Nikravesh-2006-FE}.

Currently, static features~\cite{Nikravesh-2006-FE} and dynamic features~\cite{Hadid-2011-AFB} are mainly used to identify or classify different objects in images as shown in Table~\ref{tab: A comparison table of static and dynamic features}.

\begin{table}[H]
\centering
\scriptsize 
\caption{A comparison table of static and dynamic features.}
\begin{tabular}{|c|c|l|c|l|c|l|}
\hline
\textbf{Two objects}         & \multicolumn{2}{c|}{\textbf{Static features similarity}} & \multicolumn{2}{c|}{\textbf{Dynamic features similarity}} & \multicolumn{2}{c|}{\textbf{Whether existing studies solve}} \\ \hline
\textbf{A and B}             & \multicolumn{2}{c|}{\textbf{low}}                        & \multicolumn{2}{c|}{\textbf{low}}                         & \multicolumn{2}{c|}{\textbf{Yes}}                            \\ \hline
\textbf{C and D}             & \multicolumn{2}{c|}{\textbf{low}}                        & \multicolumn{2}{c|}{\textbf{high}}                        & \multicolumn{2}{c|}{\textbf{Yes}}                            \\ \hline
\textbf{E and F}             & \multicolumn{2}{c|}{\textbf{high}}                       & \multicolumn{2}{c|}{\textbf{low}}                         & \multicolumn{2}{c|}{\textbf{Yes}}                            \\ \hline
\textbf{G and H} & \multicolumn{2}{c|}{\textbf{high}}                       & \multicolumn{2}{c|}{\textbf{high}}                        & \multicolumn{2}{c|}{\textbf{Our work}}                       \\ \hline
\end{tabular}
\label{tab: A comparison table of static and dynamic features}
\end{table}

From Table~\ref{tab: A comparison table of static and dynamic features} we can see that when facing the following three conditions, it is easy to describe the objects with existing static and dynamic features: (1) the distinction between static features and dynamic features is both high. (2) there are obvious differences in static features and little differences in dynamic features. (3) the difference between static features is little, while the difference between dynamic features is large. However, objects in microscopic videos are hard to identify or classify in the following two cases: (1) two objects with very similar static and dynamic features. (2) two objects with very weak static features and very similar dynamic features. To this end, we propose new foldover features to describe the behavior of objects in microscopic videos. 

In microscopic videos, the following difficulties usually exist in identifying or classifying different individuals of the same class of tiny objects. Firstly, because most of the tiny objects are colorless or transparent, they have little color or texture information. Secondly, when tiny objects have similar morphological characteristics, it is difficult to distinguish them by shape features. Thirdly, if the size of the objects are only several pixels, it is very hard to obtain available information. Fourthly, if two objects have both similar static and dynamic features, it is hard to identify or classify them. Hence, we select the sperm microscopic videos as the experimental material, where sperms have little color information, weak shape information, very small sizes, and similar static and dynamic features. 

\begin{figure*}[htbp]
  \centering
  \includegraphics[scale=0.31]{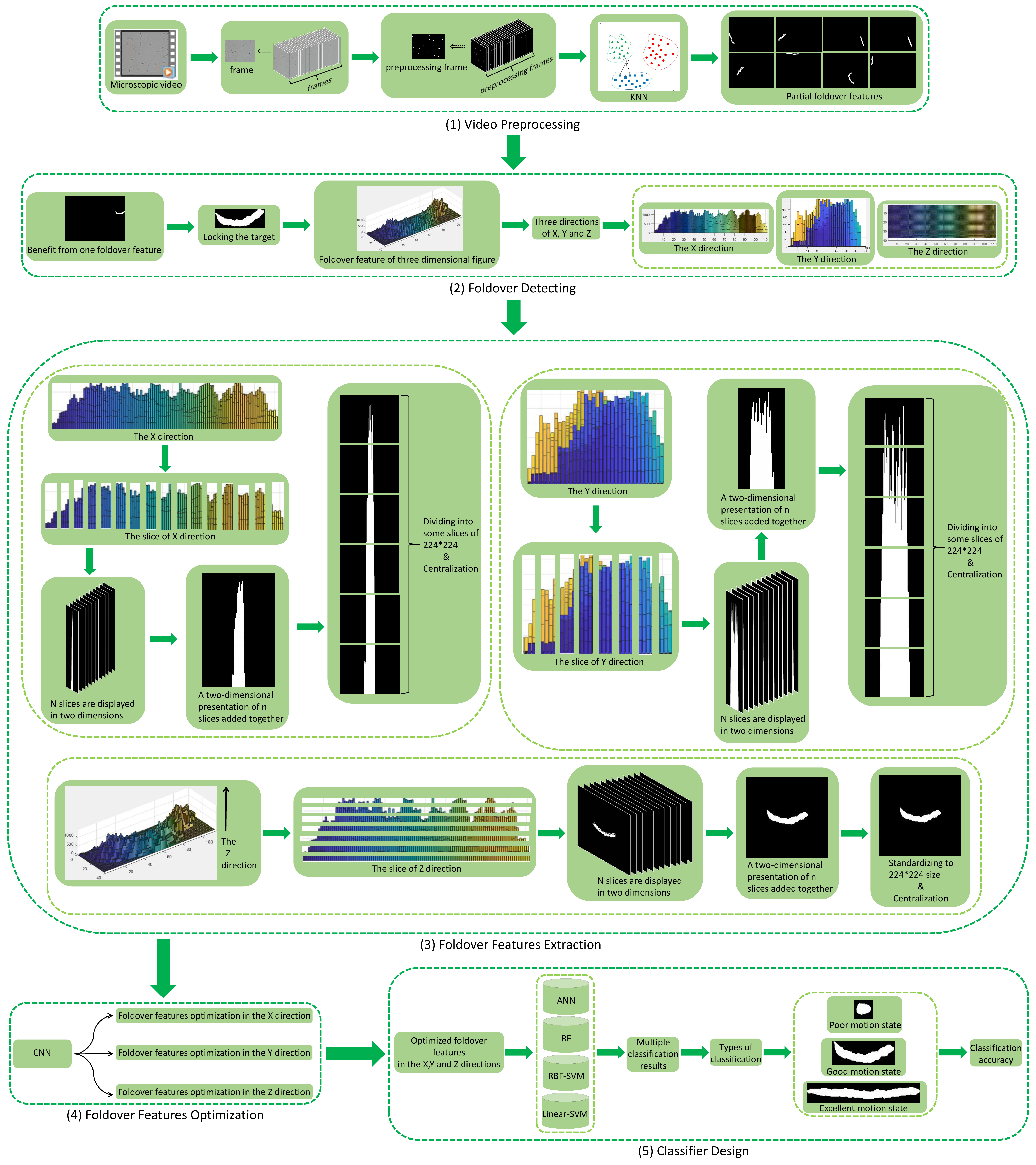}
  \caption{Work flow of the proposed foldover feature extraction method.}
  \label{fig: the_main_figure}
\end{figure*}

Foldover features are a kind of behavior feature for dynamic targets, and the work flow is shown in \figurename~\ref{fig: the_main_figure}.

There are five steps in \figurename~\ref{fig: the_main_figure}:  (1) Video preprocessing, the purpose of which is to obtain the motion foldover of each tiny object in video. (2) Foldover detection, which is used to detect the foldover of each object. (3) Foldover features extraction, extracting the foldover features from the X, Y, Z, three directions. (4) Foldover features optimization, the Convolutional Neural Network~\cite{Krizhevsky-2012-ImageNet} (CNN) removes the redundant information in the foldover features and further enhances the foldover features. (5) Classifier design, four different classifiers are used to verify the superiority of foldover features.

\section{Related work}
\label{s:classical}

This section summarizes the related works of these: \ref{s:Static Features} is about the static feature extraction methods, including various classical feature extraction methods and deep learning feature extraction methods. \ref{s:Dynamic Features} is about dynamic feature extraction methods, including several common dynamic feature extraction methods and deep learning feature extraction methods.
\subsection{Static Features}
\label{s:Static Features}
Static features usually include color, texture and shape features. Color features describe the surface properties of the scene corresponding to an image region based on the information of pixels~\cite{Long-2003-FOC}. However, when images have little color information available (such as sperm microscopic videos), the color features are almost the same. For example, in the article~\cite{Brunelli-2002-OTS}, brightness histogram is used to retrieve images with good results. However, when multiple objects have similar color brightness, this method is easy to loses effectiveness.

Texture features reflect the properties of surface structure organization and arrangement with slow change or periodic change~\cite{Deputy-1973-texturalfeatures}. For example, the proposal of the Histogram of Oriented Gradient (HOG) feature~\cite{NavneetDalal-2005-HOG}, the advantage of HOG is that the geometric deformation and optical deformation of image have little influence on HOG. However, HOG is difficult to deal with the occlusion, for example, in the sperm microscopic videos, when two sperms collide and overlap, the extraction result of HOG feature will have errors. Another example, the application of Gray-Level Co-occurrence Matrix (GLCM)~\cite{Seongjin-2011-GGO}. GLCM is used to calculate uniformity and strength values to identify candidate areas of Ground Glass Opacity (GGO) nodules. However, GLCM cannot identify two very similar objects by only describing the gray relationship between a certain pixel and a pixel within a certain distance.

There are many effective shape features, such as geometric features, Hu moments~\cite{Hu-1962-Visual}, shape signature~\cite{Giannarou-2007-Shape} and Scale-invariant Feature Transform (SIFT) features~\cite{Mortensen-2005-SIFT}. The geometric features mainly include perimeter, area, long axis, short axis, length-width ratio and complexity, which can be used for motion analysis. However, in the analysis and recognition of similar targets (such as sperm videos), it is not effective to use geometric features. Hu moments are higher order geometric features, which are used to reflect the distribution of random variables in statistics. Translation, scale expansion, rotation these changes will not affect the invariant moment, invariant moment has a good invariance. However, Hu moments depend on image segmentation a lot, and their application fields are limited. SIFT feature~\cite{Mortensen-2005-SIFT} is a local feature of images, which is invariant to rotation, scale scaling and brightness change, and has good stability to angle change, affine transformation and noise influence. However, because detection of key points is an important step in SIFT feature extraction, SIFT features extracted from tiny targets are limited. Shape signature is a boundary - based shape descriptor formed by a set of one-dimensional signals called shape signatures~\cite{Giannarou-2007-Shape}, which is robust to environmental conditions (partial occlusion) and image transformation (scaling, rotation, translation). But, the point of shape signature is to identify objects based on their shape, which is not effective at recognizing object (such as sperm) with similar shape.

With the development of deep learning technology, we can adopt different neural network frameworks to extract the deep learning features of the target objects. $Convolutional$ $Neural$ $Network$~\cite{Krizhevsky-2012-ImageNet} (CNN) is an efficient identification method, and because this network avoids the complicated pre-processing of images and can directly input the original images, and be widely used. VGG16~\cite{Simonyan-2014-Very} network is a classical CNN, which explores the relationship between the depth of the convolutional neural network and its performance, the error rate is greatly reduced and the performance is improved by deepening the network structure. For example, we can use VGG16~\cite{Simonyan-2014-Very} network to directly extract the deep learning features of static targets. Deep learning features can be used for further data statistics at the pixel level. However, when objects are tiny (such as sperms in microscopic videos ), the feature extraction ability of CNN is very limited.
\subsection{Dynamic Features}
\label{s:Dynamic Features}
With the development of pattern recognition and intelligent video processing technology, there are more and more researches on dynamic target analysis. Dynamic texture is an extension of static texture in time domain, which includes both static and dynamic information~\cite{Soatto-2001-DT}. For example, in the Motion Energy Model~\cite{ADELSON-1985-Spatiotemporal}, a video sequence is regarded as the direction in the three-dimensional space-time, and a directional selective filter is used to extract the motion information of each position. For another example, based on the expansion of separable guided filtering theory~\cite{Derpanis-2010-Dynamic}, a 3D filter is decomposed into three independent one-dimensional filters, which are filtering along the horizontal, vertical and time directions, and the filtering efficiency is significantly improved. Another example, the Gaussian Mixture Model (GMM)~\cite{Zivkovic-2004-Improved}, which is widely used to model the background of complex dynamic scenes, especially on the occasions of periodic movement, such as shaking branches, turbulent water, snowstorms and fountains. GMM can steadily and quickly detect suspected motion prospects. Another example, Mixtures of Dynamic Textures (MDT)~\cite{Chan-2008-Modeling} is used for video frame sequence modeling. MDT can use dynamic textures to generate a series of video sequences into specific samples, which has excellent performance in motion clustering and segmentation. The above four examples are target motion analysis in video based on dynamic texture. However, the basis of dynamic texture is static texture, which is an extension of static texture in the time domain. In microscopic video analysis, we encounter the following difficulties: (1) Multi-objective analysis, there are many objects of our analysis in each frame. (2) All the objects are very tiny, and there is no significant difference in the appearance of different objects (such as sperms). (3) Little texture information of tiny objects. (4) Interference of impurities, some impurities are similar to our analysis objects in appearance. The above difficulties cannot be solved by dynamic texture features.

The acquisition of motion parameters is also useful for object motion analysis. For example, a series of motion parameters of each sperm are continuously collected to analyse sperm motion~\cite{Urbano-2016-Automatic} and achieve good results. However, in the case that there are many sperm targets in the camera lens, different sperm targets have similar motion patterns and little difference in motion parameters. Therefore, it is not enough to rely on motion parameters alone.

In recent years, deep learning method has been successfully applied in object tracking field, and gradually surpasses the traditional method in performance. A typical strategy is that first obtaining the feature representation of a target by using CNNs, then the CNNs are trained on a large-scale classification database like ImageNet~\cite{Russakovsky-ImageNet-2015}, and the trained CNNs are finally used to classify and track the objects. This approach not only avoids the problem of insufficient samples of large-scaleCNN, but also makes full use of the strong representation ability of deep learning features. 

For example, FCNT mainly analyses the conv4-3 and conv5-3 output feature maps of VGG-16~\cite{Wang-2016-Visual}. FCNT constructes a feature screening network and two complementary heat-map prediction networks based on the analysis of features of different CNN layers. FCNT makes the targets more robust during deformation. For another example, the work of~\cite{Chao-2015-Hierarchical} uses the output of conv3-4, conv4-4 and conv5-4 in a pre-trained VGG-19~\cite{Simonyan-2014-Very} as the feature extraction layer. The Features extracted from these three layers are respectively studied through relevant filters to obtain different templates, and then the obtained three results are fused to obtain the final target position. However, the above method is not applicable to the identification and analysis of multi-target motion in sperm microscopic videos. The difficulties in using deep learning in the field of target tracking and recognition are appearance deformation, light change, fast movement, motion blur, interference from similar objects, scale change, occlusion and target movement out of the field of view. These difficulties are also the problems that we encounter in the sperm microscopic videos. In addition, the five difficulties proposed in this paper in the section on dynamic texture are still not well solved by using the above methods. These five difficulties are also the key problems to be solved in this paper.

\section{Foldover Features}
\label{s:Our methods}

In this section, we introduce the proposed foldover feature extraction method, referring to \ref{s:Constrction of Foldovers} foldover construction, \ref{s:Foldover Features Extraction} foldover feature extraction.

For the convenience of narration, the variables are used in this paper as follows: (1) We define a data set of videos as $\chi {\rm{=}}\left\{ {{X_1},{X_{\rm{2}}}, \ldots ,{X_i},...,{X_n}} \right\}$, $i=1, 2, 3, \ldots, n$, where $X_i$ is the video variable, $i$ is the video number, and $n$ is the total number of videos in $\chi$. Furthermore, ${X_i} = \left\{ {{x_{\left( {i,1} \right)}},{x_{\left( {i,2} \right)}},...,{x_{\left( {i,j} \right)}},...,{x_{\left( {i,m} \right)}}} \right\}$ ( $j=1, 2, 3, \ldots, m$ ) is a set of frames (static images), where $x_{\left( {i,j} \right)}$ is the frame variable, $j$ is the frame number, $m$ is the total number of frames in ${X_i}$. In addition, ${x_{\left( {i,j} \right)}} = \left\{ {{x_{\left( {i,j,1} \right)}},{x_{\left( {i,j,2} \right)}},...,{x_{\left( {i,j,k} \right)}},...,{x_{\left( {i,j,h} \right)}}} \right\}$ ( $k=1, 2, 3, \ldots, h$ ) is a set of pixels, where ${x_{\left( {i,j,k} \right)}}$ denotes the image pixel, $k$ is the pixel number, $h$ is the total number of pixels in a frame, $h = {h_1} \times {h_2}$, ${h_1=1, 2, 3, \ldots}$ is the number of pixels in a row, and ${h_2=1, 2, 3, \ldots}$ is the number of pixels in a column. (2) We define the intensity (pixel value) at pixel ${x_{\left( {i,j,k} \right)}}$ as $p\left( {{x_{\left( {i,j,k} \right)}}} \right) \in \left[ {0,255} \right]$. (3) We define a set of sperms in each frame as ${\varsigma _{\left( {i,j} \right)}} = \left\{ {{s_{\left( {i,j,1} \right)}},{s_{\left( {i,j,2} \right)}},...,{s_{\left( {i,j,l} \right)}},...,{s_{\left( {i,j,q} \right)}}} \right\}$, $l=1,2,3,...,q$, where ${{s_{\left( {i,j,l} \right)}}}$ is one sperm, $i$ is the video number, $j$ is the frame number, $l$ is the sperm number, and $q$ is the  total number of sperms in this frame. Because in the video, there are sperms constantly swimming into or out of the visual fields, therefore ${\varsigma _{\left( {i,j} \right)}}$ cannot represent the total number of sperms in the videos, and it can only represent the number of sperms in the current frame ${x_{\left( {i,j} \right)}}$.

\subsection{Constrction of Foldovers}
\label{s:Constrction of Foldovers}
There are many sperms in a semen microscopic video,  and we construct a foldover for each sperm by the following six steps.

\paragraph{\textbf{(1) Video decomposition}}

\begin{figure}[htbp]
\centering
\includegraphics[scale=0.3]{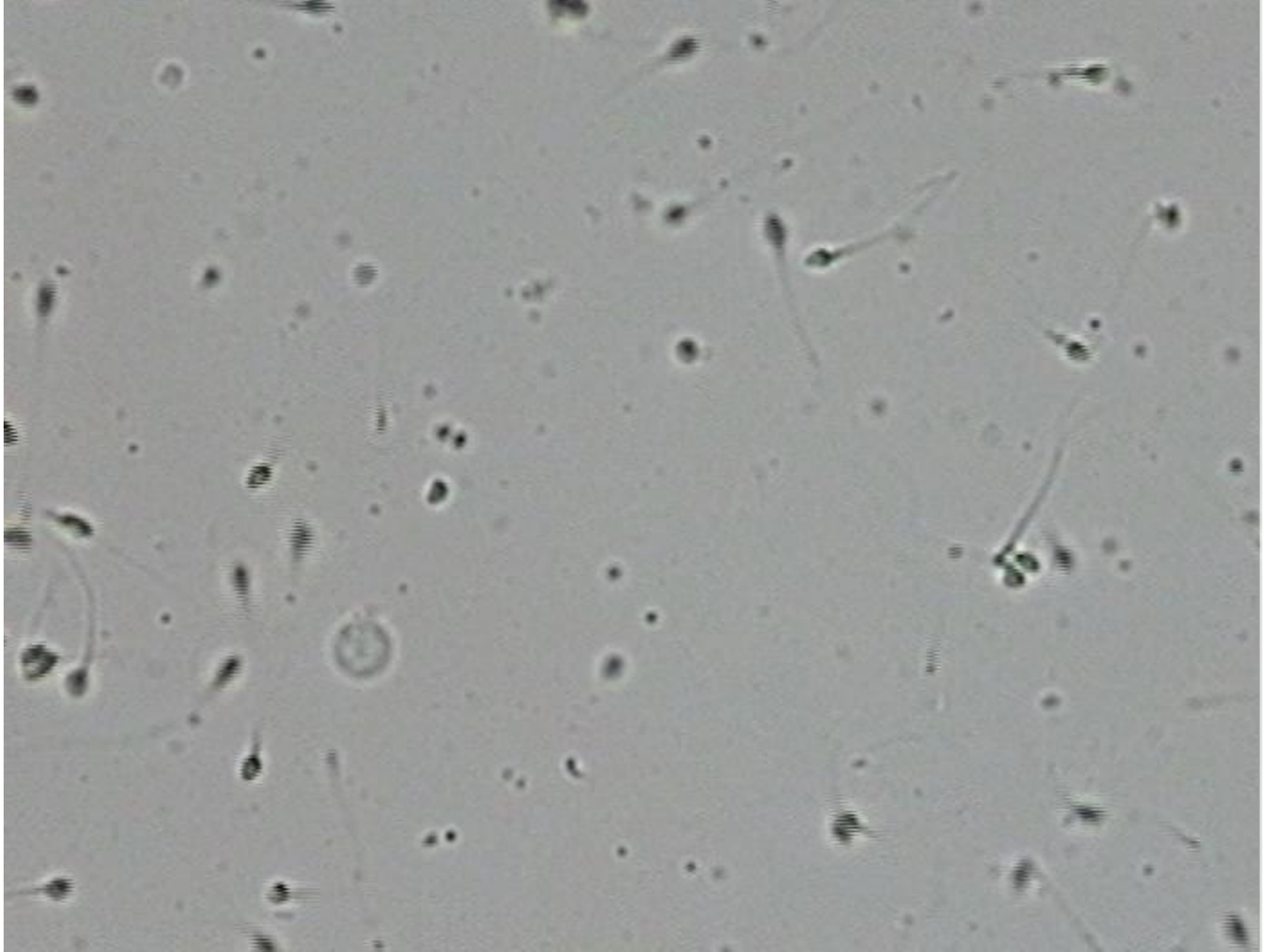}
\caption{An example of a semen microscopic video frame (a static gray-scale image) $x_{\left({i,j}\right)}$.}
\label{fig:a static gray scale image}
\end{figure}

We decompose a semen microscopic video ${X_i}$ into frames ${x_{\left( {i,1} \right)}},{x_{\left( {i,2} \right)}},...,\\{x_{\left( {i,j} \right)}},...,{x_{\left( {i,m} \right)}}$, where each frame ${x_{\left( {i,j} \right)}}$ is a static gray-scale image as shown in \figurename~\ref{fig:a static gray scale image}.

\paragraph{\textbf{(2) Image segmentation}}

We define the threshold value of the image $x_{(i,j)}$ as $T\left( {{x_{\left( {i,j} \right)}}} \right)$, the segmentation result of $x_{(i,j)}$ as ${{{x^{\rm{seg}}_{\left( {i,j} \right)}}}}$, and the value of the $k$-th pixel in $x^{\rm{seg}}_{(i,j)}$ as $p\left( {{x^{\rm{seg}}_{\left( {i,j,k} \right)}}} \right)$ in Eq.~(\ref{con: eq-1}) .

\begin{equation}
\label{con: eq-1}
p\left( {{x}^{\rm{seg}}}_{\left( i,j,k \right)} \right)=\left\{ \begin{matrix}
   0 & p\left( {{x}_{\left( i,j,k \right)}} \right)\le T\left( {{x}_{\left( i,j \right)}} \right)  \\
   1 & otherwise  \\
\end{matrix} \right.
\end{equation}

\begin{figure}[H]
\centering
\includegraphics[scale=0.3]{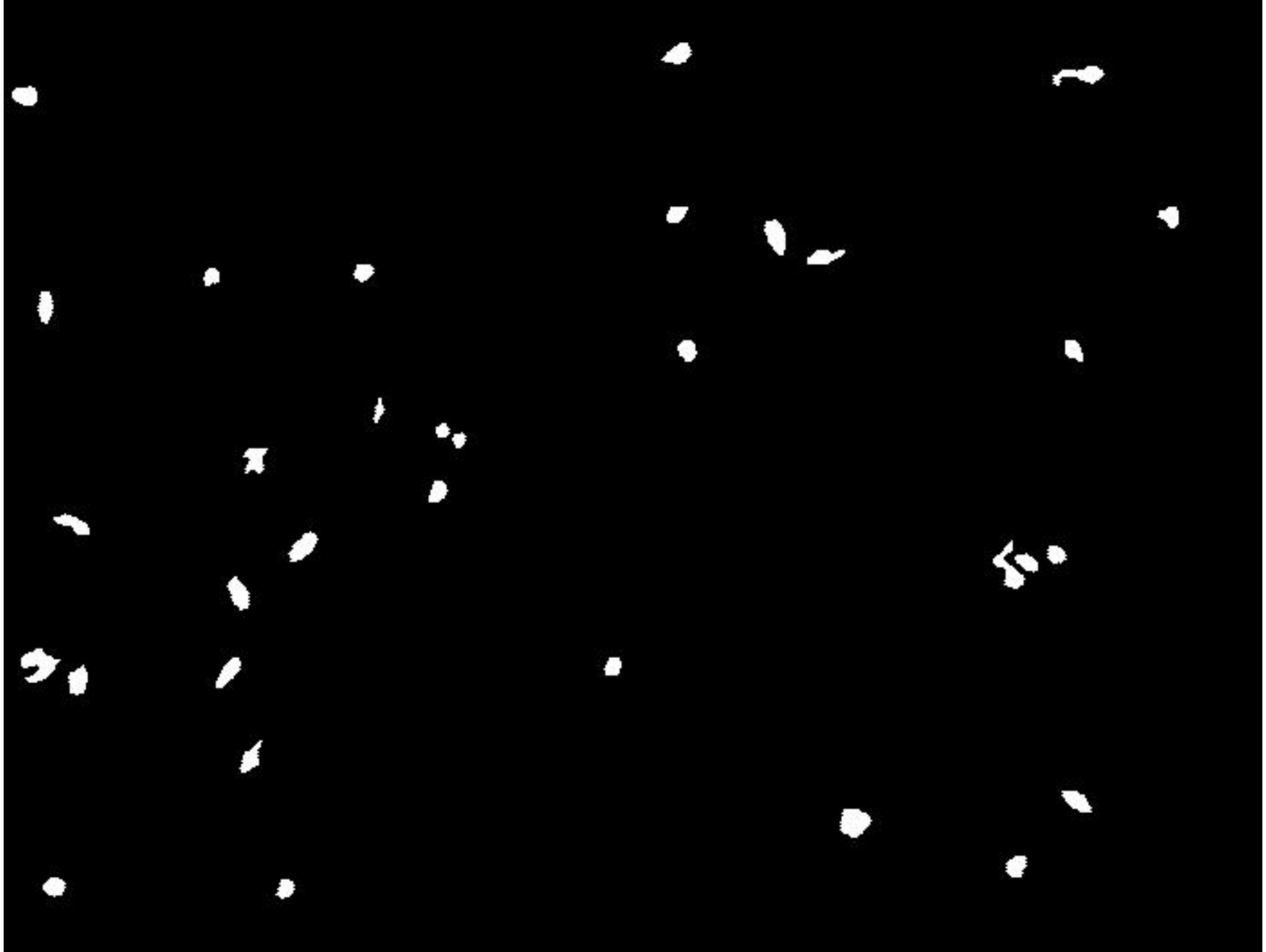}
\caption{An example of the threshold segmentation  result ${{x^{s}_{\left( {i,j} \right)}}}$.}
\label{fig:segmentation result}
\end{figure}

In Eq.~(\ref{con: eq-1}) , When the pixel value $p\left( {{x}_{\left( i,j,k \right)}} \right)$ is lower than the threshold $T\left( {{x_{\left( {i,j} \right)}}} \right)$, the result of threshold segmentation $p\left( {{x^{\rm{seg}}_{\left( {i,j,k} \right)}}} \right)$ is 0 (black); otherwise, the result of threshold segmentation $p\left( {{x^{\rm{seg}}_{\left( {i,j,k} \right)}}} \right)$ is 1 (white). Finally,   we get the image segmentation result ${{{x^{\rm{seg}}_{\left( {i,j} \right)}}}}$, and all the sperms ${\varsigma _{\left( {i,j} \right)}}$ in each frame $x_{\left( {i,j} \right)}$ are obtained. An example of the threshold segmentation  result  is shown in \figurename~\ref{fig:segmentation result}.

\paragraph{\textbf{(3) Barycenter coordinates extraction}}

Based on the image segmentation results ${{{x^{s}_{\left( {i,j} \right)}}}}$, we define a barycenter coordinates set of all sperms for total frames in the video $X_i$ as ${{\psi }_{\left( i \right)}}=\left\{ {{C}_{\left( i,1 \right)}},{{C}_{\left( i,2 \right)}},...,{{C}_{\left( i,j \right)}},...,{{C}_{\left( i,m \right)}} \right\}$, where ${{C}_{\left( i,j \right)}}$ is the barycenter coordinate variable, $i$ is the video number, $j$ is the  frame number, and $m$ is the total number of frames. Furthermore, ${{C}_{\left( i,j \right)}}=\left\{ c\left( {{s}_{\left( i,j,1 \right)}} \right),c\left( {{s}_{\left( i,j,2 \right)}} \right),\ldots, \right.$ $\left. c\left( {{s}_{\left( i,j,l \right)}} \right),...,c\left( {{s}_{\left( i,j,q \right)}} \right) \right\}$ is a set of barycenter coordinates for all sperms ${\varsigma _{\left( {i,j} \right)}}$ in the frame ${{{x_{\left( {i,j} \right)}}}}$, where $c\left( {{s}_{\left( i,j,l \right)}} \right)$ is the  barycenter coordinates of $l$-th sperm in the $j$-th frame of $i$-th video. In conclusion, we extract all barycenter coordinates ${{\psi }_{\left( i \right)}}$ from all sperms in the video $X_i$. 

\paragraph{\textbf{(4) Target matching}}

Our challenge is to match the same target from the current frame to the next frame. We choose a classical $k$-nearest neighbor ($k$-NN)~\cite{Goldstein-2003-knn} algorithm to solve this problem. The idea of $k$-NN is: In the feature space, if a sample $\varphi$ is most similar to $\kappa$ samples, and most of $\kappa$ samples belong to a certain category, then sample $\varphi$ also belongs to this category. These $\kappa$ samples are the closest to sample $\varphi$ in the feature space, where $\kappa$ is usually an integer not greater than 20. 

Based on the results of $\psi_{\left({i} \right)}$, we obtain the barycenter coordinates of all sperms ${\varsigma _{\left( {i,j} \right)}}$ in the video $X_i$. Then, we use $k$-NN algorithm to calculate Euclidean distance: There is a barycenter coordinate $c\left( {{s}_{\left( i,j,l \right)}} \right)$ in frame $x_{\left( {i,j} \right)}$, and next frame $x_{\left( {i,j+1} \right)}$, where all the barycenter coordinates are ${{C}_{\left( i,j+1 \right)}}=\left\{ c\left( {{s}_{\left( i,j+1,1 \right)}} \right),c\left( {{s}_{\left( i,j+1,2 \right)}} \right),...,c\left( {{s}_{\left( i,j+1,l \right)}} \right) \right.$ $\left. ,...,c\left( {{s}_{\left( i,j+1,q \right)}} \right) \right\}$. We calculate the Euclidean distance between $c\left( {{s}_{\left( i,j,l \right)}} \right)$ and all the barycenter coordinates in ${{C}_{\left( i,j+1 \right)}}$, and figure out a set of Euclidean distance ${D_{\left( {{\rm{i}},j} \right)}} = \left\{ {{d_{\left( {i,j,1} \right)}},{d_{\left( {i,j,2} \right)}},...,{d_{\left( {i,j,l} \right)}},...,} \right.$ $\left. {{d_{\left( {i,j,q} \right)}}} \right\}$, where ${{d}_{\left( i,j,l \right)}}$ is the Euclidean distance between $c\left( {{s}_{\left( i,j,l \right)}} \right)$ and $c\left( {{s}_{\left( i,j+1,l \right)}} \right)$ in Eq.~(\ref{con: eq-2}) .

\begin{equation}
\label{con: eq-2}
{{d}_{\left( i,j,l \right)}}=\sqrt{{{\left[ c\left( {{s}_{\left( i,j+1,l \right)}} \right)-c\left( {{s}_{\left( i,j,l \right)}} \right) \right]}^{2}}}
\end{equation}

\begin{figure}[H]
\centering
\includegraphics[scale=0.45]{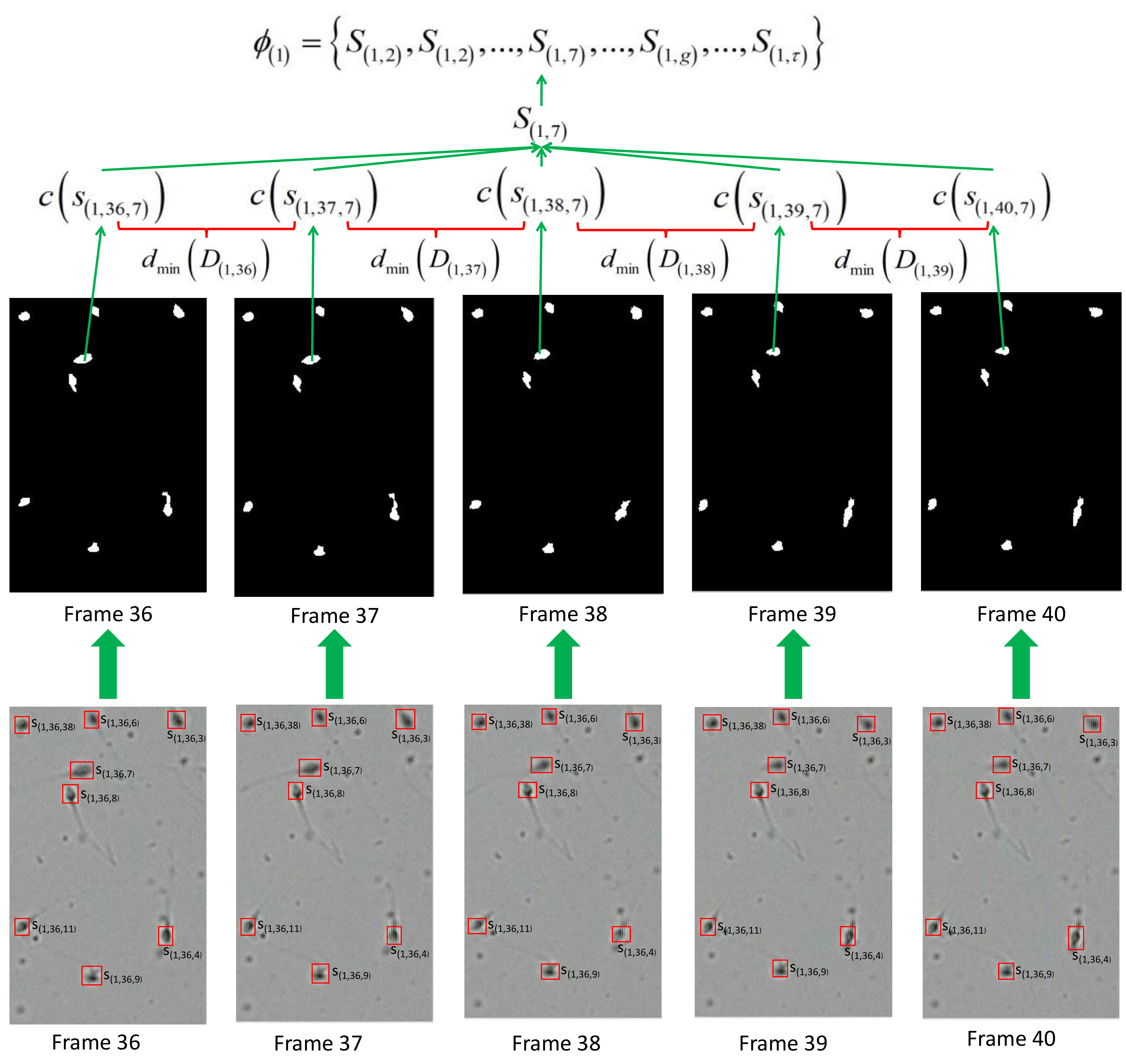}
\caption{An example of the $k$-NN classification result for sperms.}
\label{fig:KNN classification result}
\end{figure}

We find the minimum in ${{D}_{\left( i,j \right)}}$, and define this minimum as ${{d}_{\min }}\left( {{D}_{\left( i,j \right)}} \right)$. We use $k$-NN algorithm to classify all the barycentric coordinates to their corresponding coordinates in the former frame of the video. The result of classification is that all barycentric coordinates ${{\psi }_{\left( i \right)}}=\left\{ {{C}_{\left( i,1 \right)}},{{C}_{\left( i,2 \right)}},...,{{C}_{\left( i,j \right)}},...,{{C}_{\left( i,m \right)}} \right\}$ of the same sperm target in the video $X_i$ are classified into one category. An example of a classification is shown in \figurename~\ref{fig:KNN classification result}.

As the example shown in \figurename~\ref{fig:KNN classification result}, we define a set of classification result as ${{\phi }_{\left( i \right)}}=\left\{ {{S}_{\left( i,1 \right)}}, \right.$ $\left. {{S}_{\left( i,2 \right)}},...,{{S}_{\left( i,g \right)}},...,{{S}_{\left( i,\tau  \right)}} \right\}$, where ${S_{\left( {i,g} \right)}} = \left\{ {{I_{\left( {i,j,g} \right)}},} \right.$ $\left. {{I_{\left( {i,j + 1,g} \right)}},...} \right\}$ is a set of all the barycentric coordinates of one sperm in this video $X_i$, $I$ is the barycentric coordinate variable, $j$ is the frame number, $g$ is the index number of classification result, $\tau$ is the total number of classification result, and $i$ is the video number.

In the video $X_i$, there are sperms constantly swimming into or out of the visual field, therefore, sperm counts are inequality in different frames. According to this practical situation, we give a solution strategy as follow:
\begin{itemize}
\item {\bfseries Case-I: } If there is a sperm swimming into the visual field, we define this sperm as a new target, and it will have a new classification result for its own with the $k$-NN classifier. 
\item {\bfseries Case-II: } If there is a sperm swimming out of the visual field, we stipulate that the motion of this sperm is over.
\end{itemize} 

\begin{figure}[H]
\centering
\includegraphics[scale=0.3]{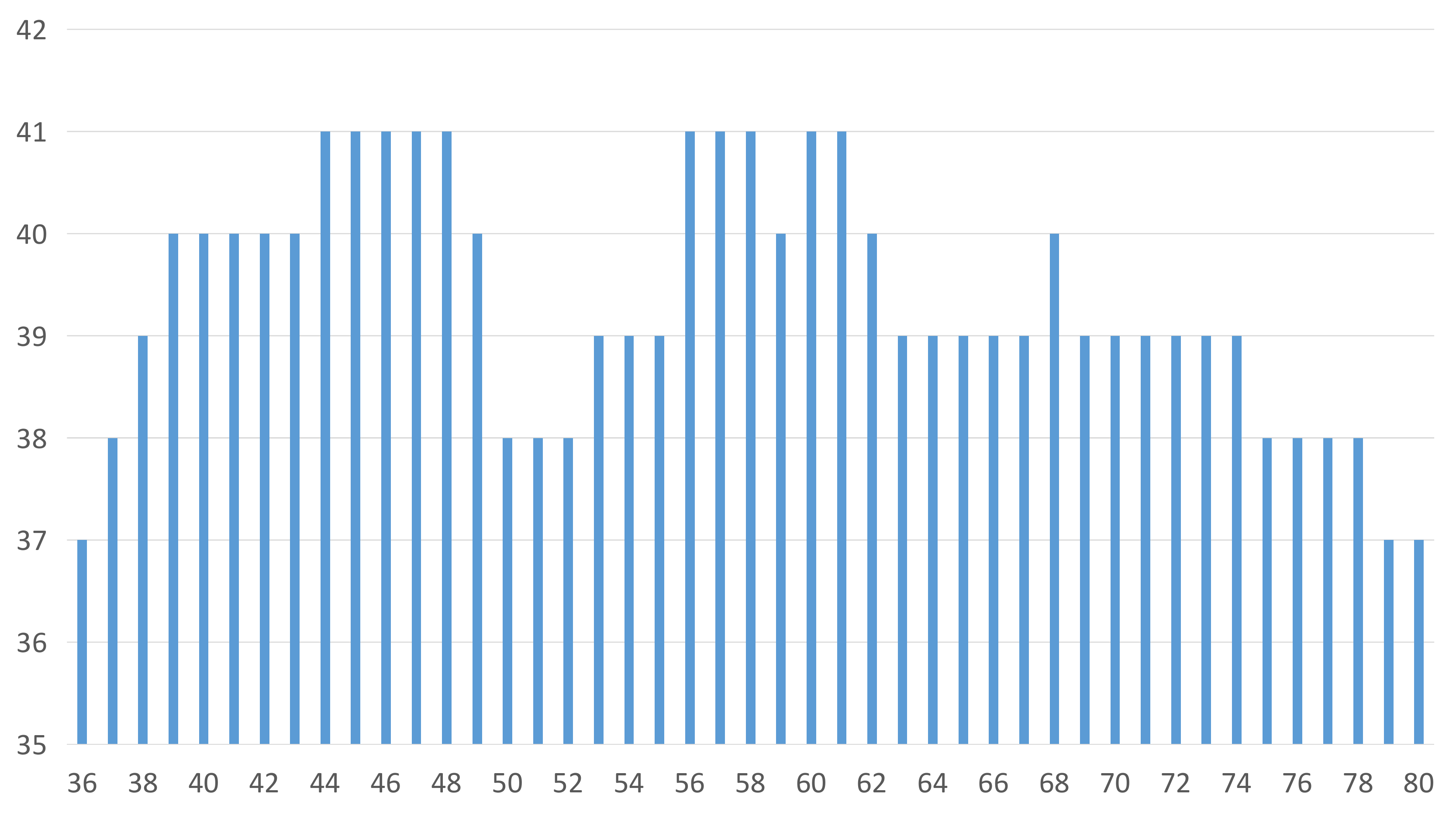}
\caption{The total number of sperms from frame 36 to frame 80 in video $X_i$.}
\label{fig:Histogram of sperm}
\end{figure}

Based on {\bfseries Case-I} and {\bfseries Case-II}, we can conclude that the number of  classification result ${\phi _{\left( i \right)}} = \left\{ {{S_{\left( {i,1} \right)}},{S_{\left( {i,2} \right)}},...,{S_{\left( {i,g} \right)}}} \right.$ $\left. {,...,{S_{\left( {i,\tau } \right)}}} \right\}$ is the total number of sperms in the video $X_i$, where $\tau$ is the total number of sperms. \figurename~\ref{fig:Histogram of sperm} shows an example of the sperm count statistics from frame 36 to frame 80 in video $X_i$. 

\paragraph{\textbf{(5) Construction of the foldover}}

According to the result of $k$-NN classification, we get the barycentric coordinates ${\phi _{\left( i \right)}} = \left\{ {{S_{\left( {i,1} \right)}},{S_{\left( {i,2} \right)}},...,{S_{\left( {i,g} \right)}}} \right.$ $\left. {,...,{S_{\left( {i,\tau } \right)}}} \right\}$ of all the sperms in the video $X_i$. The following operations are performed for each $k$-NN classification result ${\phi _{\left( i \right)}}$. First, according to $k$-NN classification result ${S_{\left( {{\rm{i}},g} \right)}}$, we determine the range of the frames in which the sperm moves. Then, we extract these frames from the segmentation results $X_i^{{\rm{seg}}} = \left\{ {x_{\left( {i,1} \right)}^{{\rm{seg}}},x_{\left( {i,2} \right)}^{{\rm{seg}}},...,x_{\left( {i,j} \right)}^{{\rm{seg}}},...,x_{\left( {i,m} \right)}^{{\rm{seg}}}} \right\}$ according to the range of frames. In these extracted frames, setting the barycentric coordinates ${S_{\left( {i,g} \right)}} = \left\{ {{I_{\left( {i,j,g} \right)}},{I_{\left( {i,j + 1,g} \right)}},...} \right\}$ as the center, setting $r$ pixels as a standard radius. We calculate the distance between the barycentric coordinates ${S_{\left( {i,g} \right)}} = \left\{ {{I_{\left( {i,j,g} \right)}},{I_{\left( {i,j + 1,g} \right)}},...} \right\}$ and all other pixels in the $X_i^{{\rm{seg}}} = \left\{ {x_{\left( {i,1} \right)}^{{\rm{seg}}},x_{\left( {i,2} \right)}^{{\rm{seg}}},...,x_{\left( {i,j} \right)}^{{\rm{seg}}},...,x_{\left( {i,m} \right)}^{{\rm{seg}}}} \right\}$, so the pixel value $p\left[ {{L_{\left( {i,j,g} \right)}}\left( {x_{\left( {i,j,k} \right)}^{{\rm{seg}}}} \right)} \right]$ is defined as Eq.~(\ref{con: eq-3}) .

\begin{equation}
\label{con: eq-3}
\begin{split}
p\left[ {{L}_{\left( i,j,g \right)}}\left( x_{\left( i,j,k \right)}^{\text{seg}} \right) \right]=\left\{ \begin{matrix}
   p\left( x_{\left( i,j,k \right)}^{\text{seg}} \right) & \sqrt{{{\left( x_{\left( i,j,k \right)}^{\text{seg}}-{{I}_{\left( i,j,g \right)}} \right)}^{2}}}\le r  \\
   0 & otherwise  \\
\end{matrix} \right.
\end{split}
\end{equation}

\begin{figure}[H]
\centering
\includegraphics[scale=0.3]{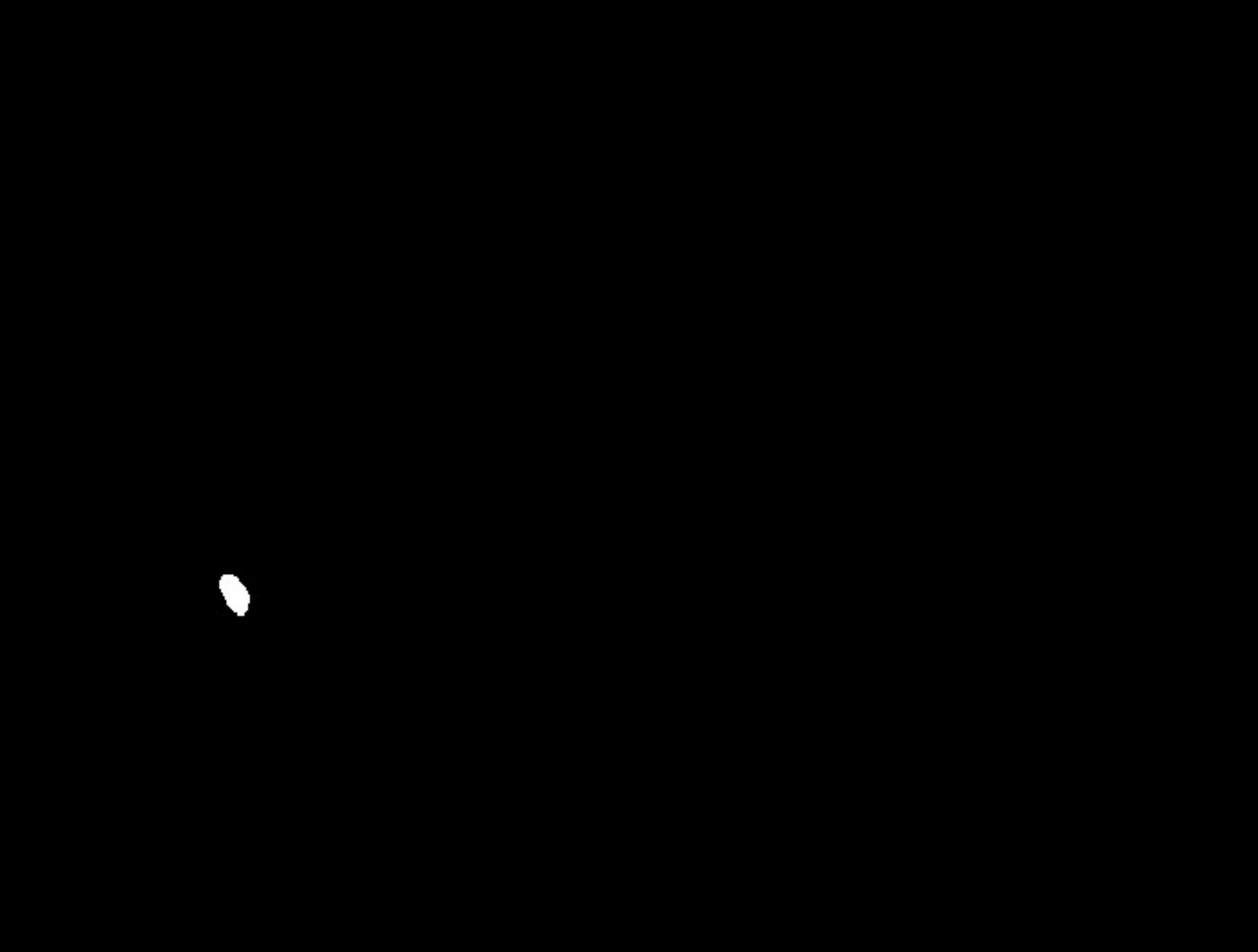}
\caption{An example of a ${{L_{\left( {i,j,g} \right)}}\left( {x_{\left( {i,j} \right)}^{{\rm{seg}}}} \right)}$ result.}
\label{fig:Sperm lock}
\end{figure}

The pixel value $p\left[ {{L_{\left( {i,j,g} \right)}}\left( {x_{\left( {i,j,k} \right)}^{{\rm{seg}}}} \right)} \right]$ is 0 which is more than $r$ pixels away from the  barycentric coordinate ${{I_{\left( {i,j,g} \right)}}}$; otherwise, the pixel value $p\left[ {{L_{\left( {i,j,g} \right)}}\left( {x_{\left( {i,j,k} \right)}^{{\rm{seg}}}} \right)} \right]$ is ${p\left( {x_{\left( {i,j,k} \right)}^{{\rm{seg}}}} \right)}$. In this way, we target each sperm in the segmentation results, and we define the result as ${{L_{\left( {i,j,g} \right)}}\left( {x_{\left( {i,j} \right)}^{{\rm{seg}}}} \right)}$, an example of the ${{L_{\left( {i,j,g} \right)}}\left( {x_{\left( {i,j} \right)}^{{\rm{seg}}}} \right)}$ result is shown in \figurename~\ref{fig:Sperm lock}.

Second, according to the ${{L_{\left( {i,j,g} \right)}}\left( {x_{\left( {i,j} \right)}^{{\rm{seg}}}} \right)}$ result, we can get a set ${{\theta }_{\left( i,g \right)}}=\left\{ {{L}_{\left( i,j,g \right)}}\left( x_{\left( i,j \right)}^{\text{seg}} \right), \right.$ $\left. {{L}_{\left( i,j+1,g \right)}}\left( x_{\left( i,j+1 \right)}^{\text{seg}} \right),... \right\}$ of the same sperm. We use ${{L_{\left( {i,j,g} \right)}}\left( {x_{\left( {i,j} \right)}^{{\rm{seg}}}} \right)}$ to localize the sperm region from the original frame (image) according to Eq.~(\ref{con: eq-4}) .

\begin{equation}
\label{con: eq-4}
p\left[ {{o_{\left( {i,j,g} \right)}}\left( {{x_{\left( {i,j,k} \right)}}} \right)} \right]\!=\!\left\{ {\begin{array}{*{20}{c}}
0&\!{p\left[ {{L_{\left( {i,j,g} \right)}}\left( {x_{\left( {i,j,k} \right)}^{{\rm{seg}}}} \right)} \right]\!=\!0}\\
{p\left( {{x_{\left( {i,j,k} \right)}}} \right)}&\!{otherwise}
\end{array}} \right.
\end{equation}

\begin{figure}[H]
\centering
\includegraphics[scale=0.3]{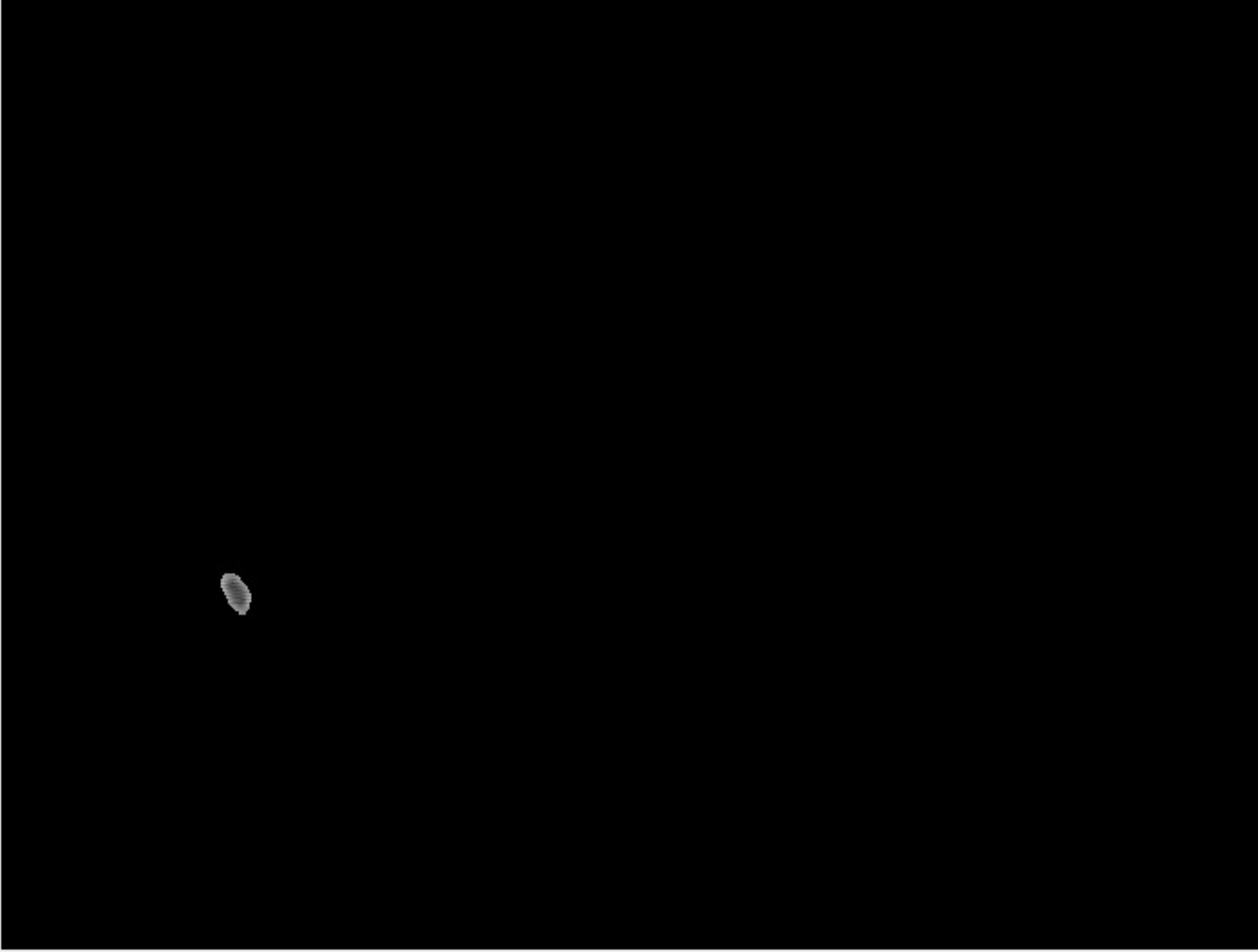}
\caption{An example of a ${{o_{\left( {i,j,g} \right)}}\left( {{x_{\left( {i,j} \right)}}} \right)}$ result.}
\label{fig:original sperm lock}
\end{figure}

In Eq.~(\ref{con: eq-4}) , we define the extracted result as ${{o_{\left( {i,j,g} \right)}}\left( {{x_{\left( {i,j} \right)}}} \right)}$. If the pixel value ${p\left[ {{L_{\left( {i,j,g} \right)}}\left( {x_{\left( {i,j,k} \right)}^{{\rm{seg}}}} \right)} \right]}$ is equal to 0, the pixel value $p\left[ {{o_{\left( {i,j,g} \right)}}\left( {{x_{\left( {i,j,k} \right)}}} \right)} \right]$ is 0; otherwise, the pixel value $p\left[ {{o_{\left( {i,j,g} \right)}}\left( {{x_{\left( {i,j,k} \right)}}} \right)} \right]$ is the pixel value ${p\left( {{x_{\left( {i,j,k} \right)}}} \right)}$ corresponding to the original image. \figurename~\ref{fig:original sperm lock} is an example of ${{o_{\left( {i,j,g} \right)}}\left( {{x_{\left( {i,j} \right)}}} \right)}$.

Thirdly, a set of the ${{o_{\left( {i,j,g} \right)}}\left( {{x_{\left( {i,j} \right)}}} \right)}$ is denoted as ${O_{\left( {i,g} \right)}}$, where we define ${\Gamma _{\left( {i,g} \right)}}\left( {{O_{\left( {i,g} \right)}}} \right)$ as the total number of extracted results in ${O_{\left( {i,g} \right)}}$, and ${O_{\left( {i,g} \right)}}$ is defined as Eq.~(\ref{con: eq-5}) .

\begin{equation}
\begin{aligned}
\label{con: eq-5}
{{O}_{\left( i,g \right)}}=\left\{ {{o}_{\left( i,j,g \right)}}\left( {{x}_{\left( i,j \right)}} \right),{{o}_{\left( i,j+1,g \right)}}\left( {{x}_{\left( i,j+1 \right)}} \right),..., \right.\ \left. {{o}_{\left( i,{{\Gamma }_{\left( i,g \right)}}\left( {{O}_{\left( i,g \right)}} \right),g \right)}}\left( {{x}_{\left( i,{{\Gamma }_{\left( i,g \right)}}\left( {{O}_{\left( i,g \right)}} \right) \right)}} \right) \right\}
\end{aligned}
\end{equation}

Based on the definitions above, we define ${\digamma _{\left( {i,g} \right)}}$ as the foldover of the $g$-th sperm in the $i$-th video, and $p\left[ {{\digamma_{\left( {i,g} \right)}}\left( {{x_{\left( {i,j,k} \right)}}} \right)} \right]$ is expressed by Eq.~(\ref{con: eq-6}) .

\begin{equation}
\label{con: eq-6}
p\left[ {{\digamma_{\left( {i,g} \right)}}\left( {{x_{\left( {i,j,k} \right)}}} \right)} \right]{\rm{ = }}\sum\limits_j^{{\Gamma _{\left( {i,g} \right)}}\left( {{O_{\left( {i,g} \right)}}} \right)} {p\left[ {{o_{\left( {i,j,g} \right)}}\left( {{x_{\left( {i,j,k} \right)}}} \right)} \right]} 
\end{equation}

\begin{figure}[H]
\centering
\includegraphics[scale=0.5]{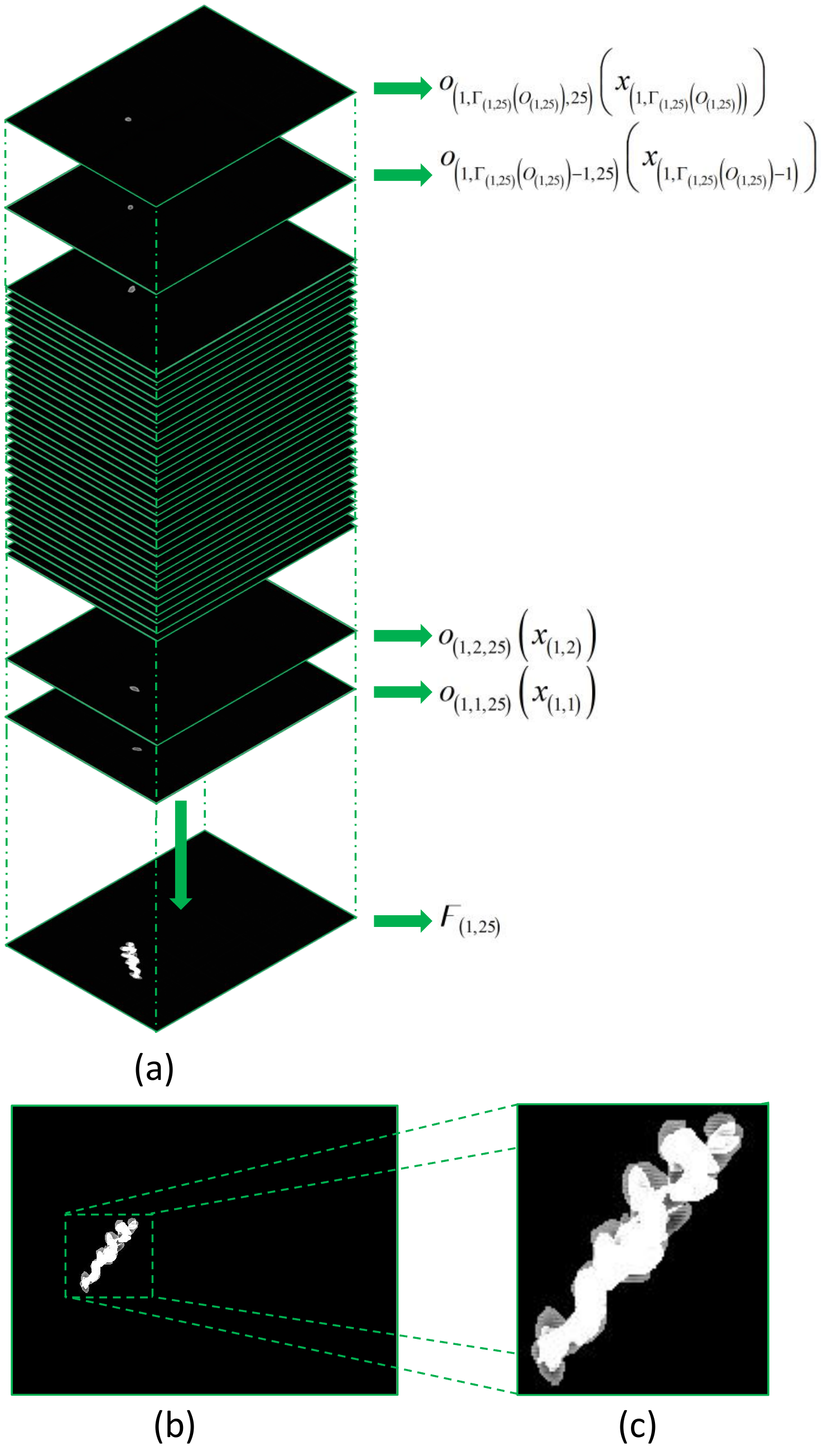}
\caption{An example of foldover ${\digamma _{\left( {i,g} \right)}}$ construction. (a) is the process of ${O_{\left( {i,g} \right)}}$ accumulation. We add up corresponding pixels in ${O_{\left( {i,g} \right)}}$, and the cumulative result is the foldover ${\digamma _{\left( {i,g} \right)}}$. (b) is the cumulative result of the ${O_{\left( {i,g} \right)}}$. (c) is the two-dimensional visualization of the foldover ${\digamma _{\left( {i,g} \right)}}$.}
\label{fig:foldover}
\end{figure}

As the definition in Eq.~(\ref{con: eq-6}) , we add up the $k$-th pixel of each frame in ${O_{\left( {i,g} \right)}}$, and the sum is $p\left[ {{\digamma_{\left( {i,g} \right)}}\left( {{x_{\left( {i,j,k} \right)}}} \right)} \right]$, $k=1, 2, 3, \ldots, h$, $k$ is the pixel number, $h$ is the total number of pixels in a frame, $h = {h_1} \times {h_2}$, ${h_1=1, 2, 3, \ldots}$ is the number of pixels in a row, and ${h_2=1, 2, 3, \ldots}$ is the number of pixels in a column. In this way, we add the corresponding pixel values in different frames to obtain ${\digamma _{\left( {i,g} \right)}}$, as shown in \figurename~\ref{fig:foldover}.

\begin{figure}[H]
\centerline{\includegraphics[scale=0.8]{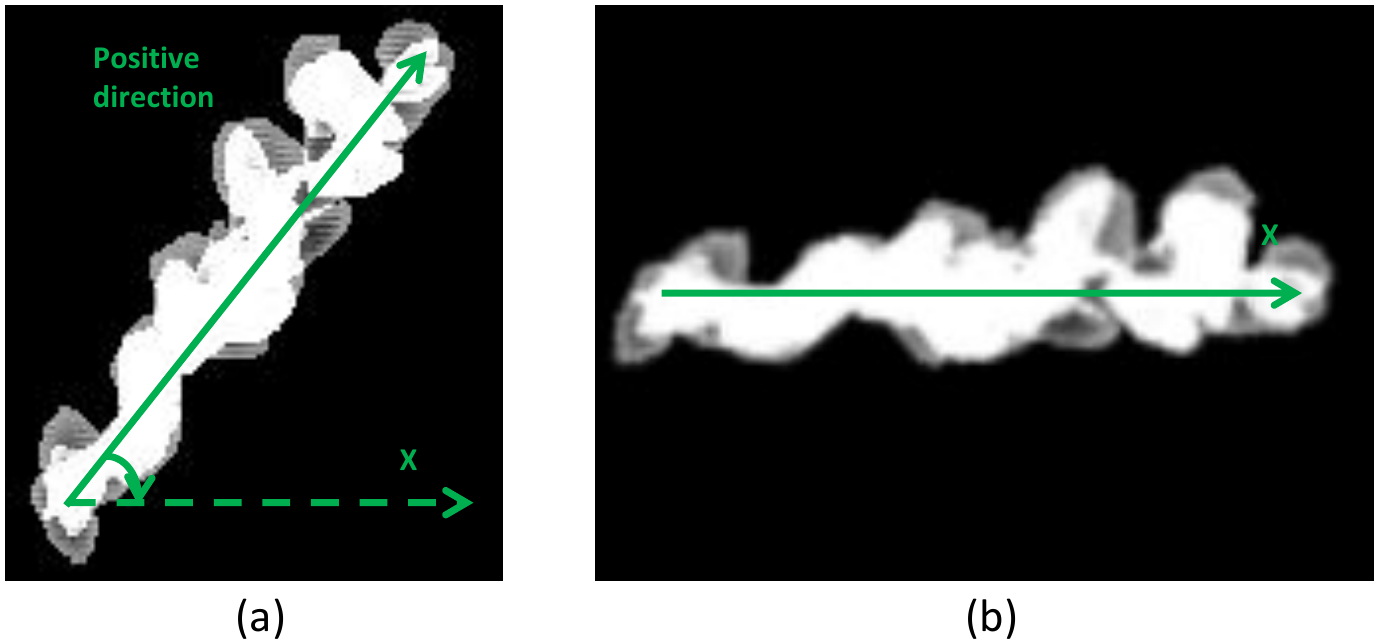}}
\caption{An example of $\digamma_{\left( {i,g} \right)}^R$ in 2D vision. (a) is the foldover ${\digamma _{\left( {i,g} \right)}}$ before the rotation. (b) is the rotated foldover $\digamma_{\left( {i,g} \right)}^R$.}
\label{fig:foldover rotate}
\end{figure}

\paragraph{\textbf{(6) Construction of 3D images}}
In the video $X_i$, the swimming directions of sperms are uncertain. Therefore, we need to unify the swimming directions of sperms to facilitate our experimental analysis. We define the direction in which the starting barycentric coordinate of sperms to their ending coordinate as the positive direction (forward direction), and the horizontal direction is defined as the X direction. In order to unify the swimming directions, we rotate the foldover ${\digamma _{\left( {i,g} \right)}}$ into this positive direction to the X direction, and we define the rotated foldover ${\digamma _{\left( {i,g} \right)}}$ as $\digamma_{\left( {i,g} \right)}^R$. An example of $\digamma_{\left( {i,g} \right)}^R$ is shown in \figurename~\ref{fig:foldover rotate}.

\begin{figure}[H]
\centerline{\includegraphics[scale=0.3]{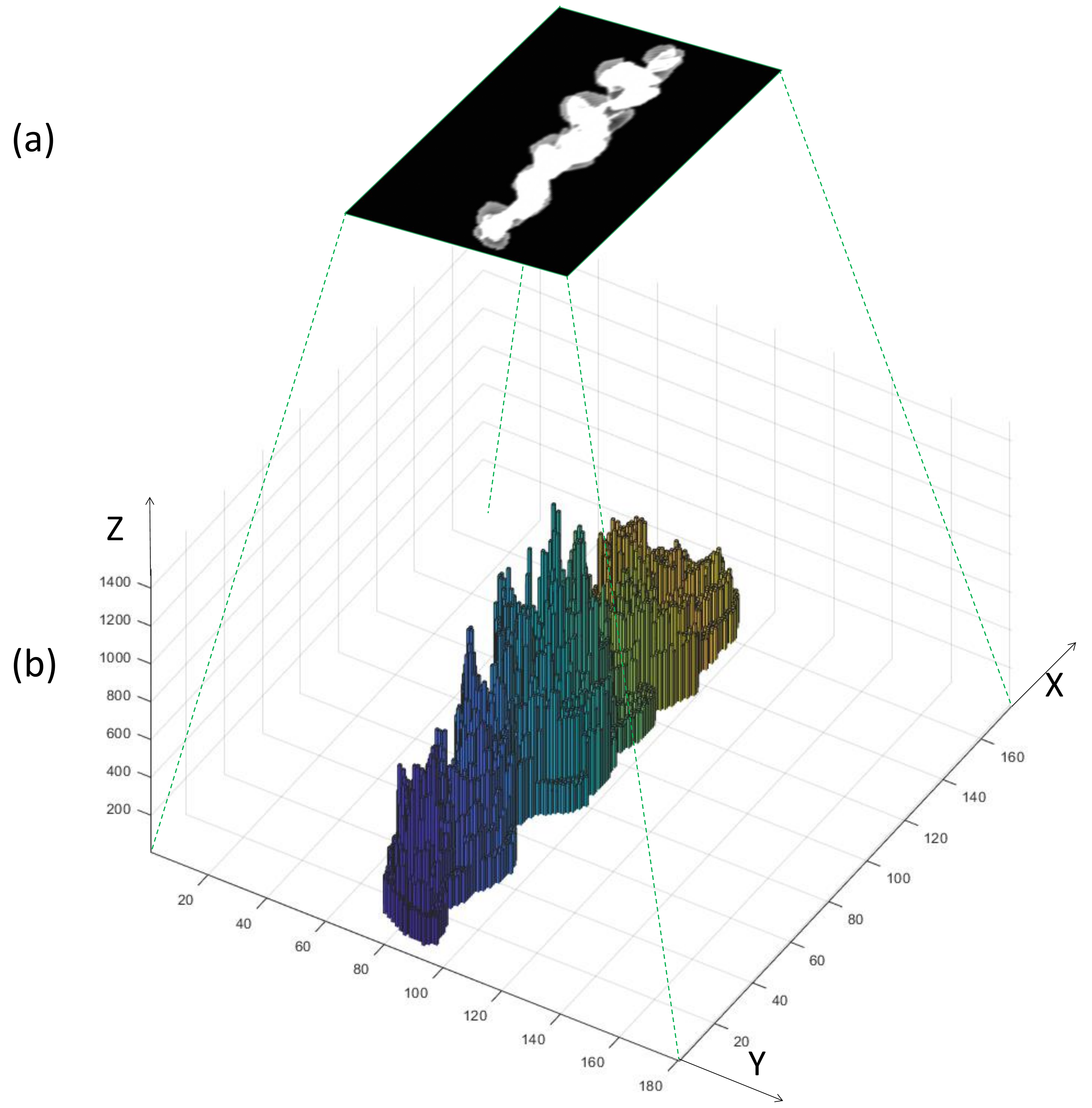}}
\caption{An example of the $\digamma_{\left( {i,g} \right)}^R$ in 3D vision. (a) is the two-dimensional visualization of the foldover $\digamma_{\left( {i,g} \right)}^R$. (b) is a 3D vision of the foldover $\digamma_{\left( {i,g} \right)}^R$.}
\label{fig:3D foldover}
\end{figure}

\figurename~\ref{fig:foldover rotate} is only the two-dimensional visualization result of the $\digamma_{\left( {i,g} \right)}^R$, it cannot contain all the information of the $\digamma_{\left( {i,g} \right)}^R$. Therefore, we show a 3D vision of the $\digamma_{\left( {i,g} \right)}^R$ to reflect all the information of the foldover in \figurename~\ref{fig:3D foldover}.

\subsection{Foldover Features Extraction}
\label{s:Foldover Features Extraction}
Foldover feature extraction is the statistics of the information in the $\digamma_{\left( {i,g} \right)}^{R}$, and the method of foldover feature extraction consists of the following four steps.

\paragraph{\textbf{(1) Foldover processing in the X, Y, and Z directions}}
First, we define the length of $\digamma_{\left( {i,g}\right)}^{R}$ on X, Y and Z three directions as $\digamma_{\left( {i,g} \right)}^R\left( \Theta  \right)$, where $\Theta$ is defined in Eq.~(\ref{con: eq-7}) .

\begin{equation}
\label{con: eq-7}
\Theta {\rm{ = }}\left\{ {\begin{array}{*{20}{c}}
\mbox{X}&{{\mbox{along the X axis}}}\\
\mbox{Y}&{{\mbox{along the Y axis}}}\\
\mbox{Z}&{{\mbox{along the Z axis}}}
\end{array}} \right.
\end{equation}

Second, we cut the foldover $\digamma_{\left( {i,g} \right)}^{R}$ along the direction of $\Theta$ with a step length of ${\nu _\Theta}$, and we can receive a set of slices which is defined as $\digamma_{\left( {i,g} \right)}^{\left( {R,\Theta} \right)}$ in Eq.~(\ref{con: eq-8}) , $u$ is the number, and ${\frac{{\digamma_{\left( {i,g} \right)}^R\left( \Theta  \right)}}{{{\nu _\Theta }}}}$ is the total number.

\begin{equation}
\label{con: eq-8}
\digamma_{\left( {i,g} \right)}^{\left( {R,\Theta} \right)}{\rm{ = }}\left\{ {\digamma_{\left( {i,g,1} \right)}^{\left( {R,\Theta} \right)},\digamma_{\left( {i,g,2} \right)}^{\left( {R,\Theta} \right)},{...,\digamma_{\left( {i,g,u} \right)}^{\left( {R,\Theta } \right)},}...,\digamma_{\left( {i,g,\frac{{\digamma_{\left( {i,g} \right)}^R\left( {\rm{\Theta}} \right)}}{{{\nu _\Theta}}}} \right)}^{\left( {R,\Theta} \right)}} \right\}
\end{equation}

Third, in X and Y directions, $\digamma_{\left( {i,g} \right)}^{R}$ can reflect time information and movement information of sperms, but $\digamma_{\left( {i,g} \right)}^{R}$ cannot reflect the information of pixel accumulation. For slices $\digamma_{\left( {i,g} \right)}^{\left( {R,\Theta} \right)}$ ($\Theta$ = X or Y), we set the pixel values of the areas where the foldover exists to 1 and other areas to 0. We add $\digamma_{\left( {i,g} \right)}^{\left( {R,\Theta} \right)}$ ($\Theta$ = X or Y) together as the result of $\digamma_{\left( {i,g} \right)}^{R}$ in the X and Y directions. Unlike the foldover slices in the X and Y directions, the foldover slices in Z direction truly reflect the effect of pixel accumulation. Therefore, we have no necessary to set the pixel values, so the pixel values of the areas where the foldover slices in the Z direction exists are added directly. We define the cumulative result of foldover slices $\digamma_{\left( {i,g} \right)}^{\left( {R,\Theta} \right)}$ as $U\left( {\digamma_{\left( {i,g} \right)}^{\left( {R,\Theta } \right)}} \right)$ ($\Theta$ = X, Y, or Z) in Eq.~(\ref{con: eq-9}) .

\begin{equation}
\label{con: eq-9}
U\left( {\digamma_{\left( {i,g} \right)}^{\left( {R,\Theta } \right)}} \right) = \sum\limits_{u = 1}^{\frac{{\digamma_{\left( {i,g} \right)}^{\left( R \right)}\left( \Theta  \right)}}{{{\nu _\Theta }}}} {\digamma_{\left( {i,g,u} \right)}^{\left( {R,\Theta } \right)}}
\end{equation}

\begin{figure}[H]
\centerline{\includegraphics[scale=0.3]{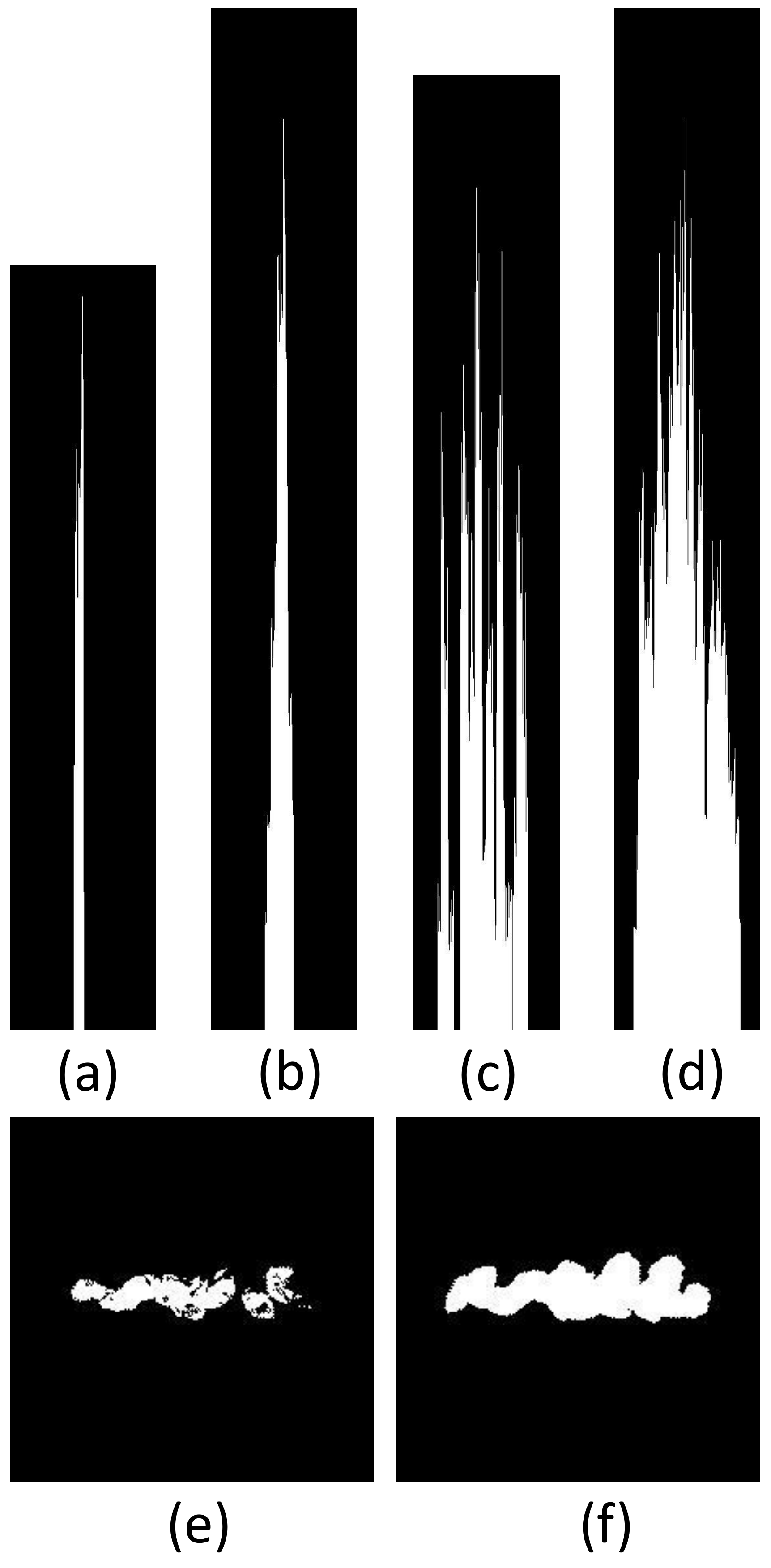}}
\caption{A slicing example and cumulative results of $U\left( {\digamma_{\left( {i,g} \right)}^{\left( {R,\Theta } \right)}} \right)$. One of the slices in X,Y and Z directions ( $\digamma_{\left( {i,g,u} \right)}^{\left( {R,X} \right)}$, $\digamma_{\left( {i,g,u} \right)}^{\left( {R,Y} \right)}$ and $\digamma_{\left( {i,g,u} \right)}^{\left( {R,Z} \right)}$ ) is shown in(a), (c) and (e). The cumulative result of slices ( $U\left( {\digamma_{\left( {i,g} \right)}^{\left( {R,X} \right)}} \right)$, $U\left( {\digamma_{\left( {i,g} \right)}^{\left( {R,Y} \right)}} \right)$ and $U\left( {\digamma_{\left( {i,g} \right)}^{\left( {R,Z} \right)}} \right)$ ) are shown in (b), (d) and (f).}
\label{fig:sperm foldoverXYZ}
\end{figure}

Finally, we get three cumulative results $U\left( {\digamma_{\left( {i,g} \right)}^{\left( {R,\Theta } \right)}} \right)$ of the foldover slices $\digamma_{\left( {i,g} \right)}^{\left( {R,\Theta} \right)}$ in  X, Y, and Z directions as shown in \figurename~\ref{fig:sperm foldoverXYZ}.

\begin{figure}[H]
\centerline{\includegraphics[scale=0.3]{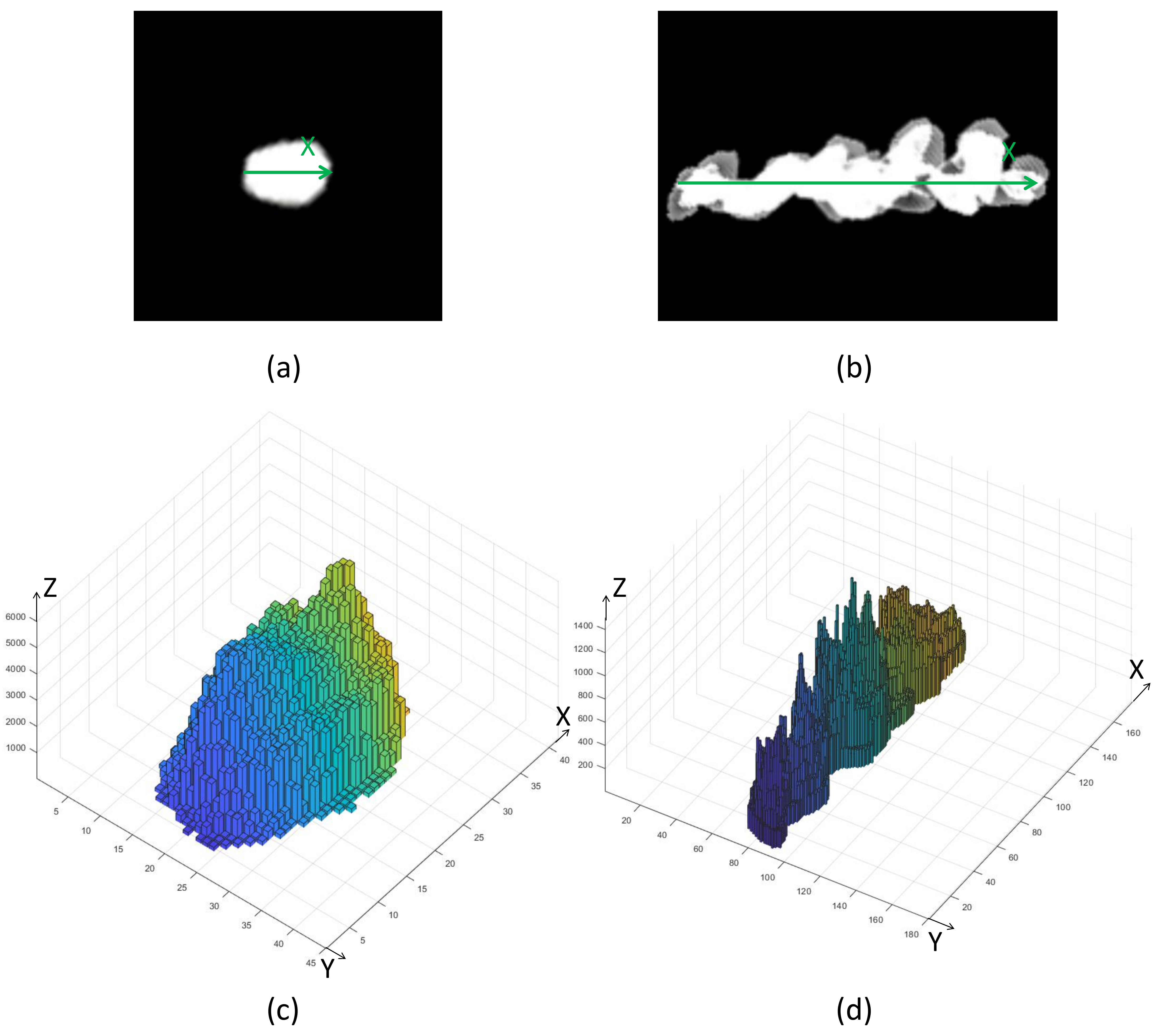}}
\caption{A comparison of foldovers of two different sperms. (a) (b) are 2D visualizations of foldovers. (c) (d) are 3D of the foldovers.}
\label{fig:different sperm foldover}
\end{figure}

\paragraph{\textbf{(2) Foldover features extraction}}
Foldovers contain different behavior information in different directions. As the example shown in \figurename~\ref{fig:different sperm foldover}, two sperms with different behaviors have completely different foldovers.

According to the human sperm quality assessment proposed by the World Health Organization (WHO) ~\cite{Lamb-WHO-2000}, sperm motility is grouped into four categories as shown in Table~\ref{tab: WHO}.

\begin{table}[H]
\centering
\normalsize  
\caption{Human sperm motility grade by World Health Organization (WHO)~\cite{Lamb-WHO-2000}}
\begin{tabular}{cccc}
\hline
\textbf{No.} & \textbf{Grades} & \textbf{Name}              & \textbf{Movements µm/s}        \\ \hline
1            & A               & Rapid progressive motility & 25                             \\ \hline
2            & B               & Slow progressive motility  & 5 \textless speed \textless 25 \\ \hline
3            & C               & non-progressive motility   & \textless 5                    \\ \hline
4            & D               & Immotility                 & No Movements                   \\ \hline
\end{tabular}
\label{tab: WHO}
\end{table}

\begin{figure}[H]
\centerline{\includegraphics[scale=0.3]{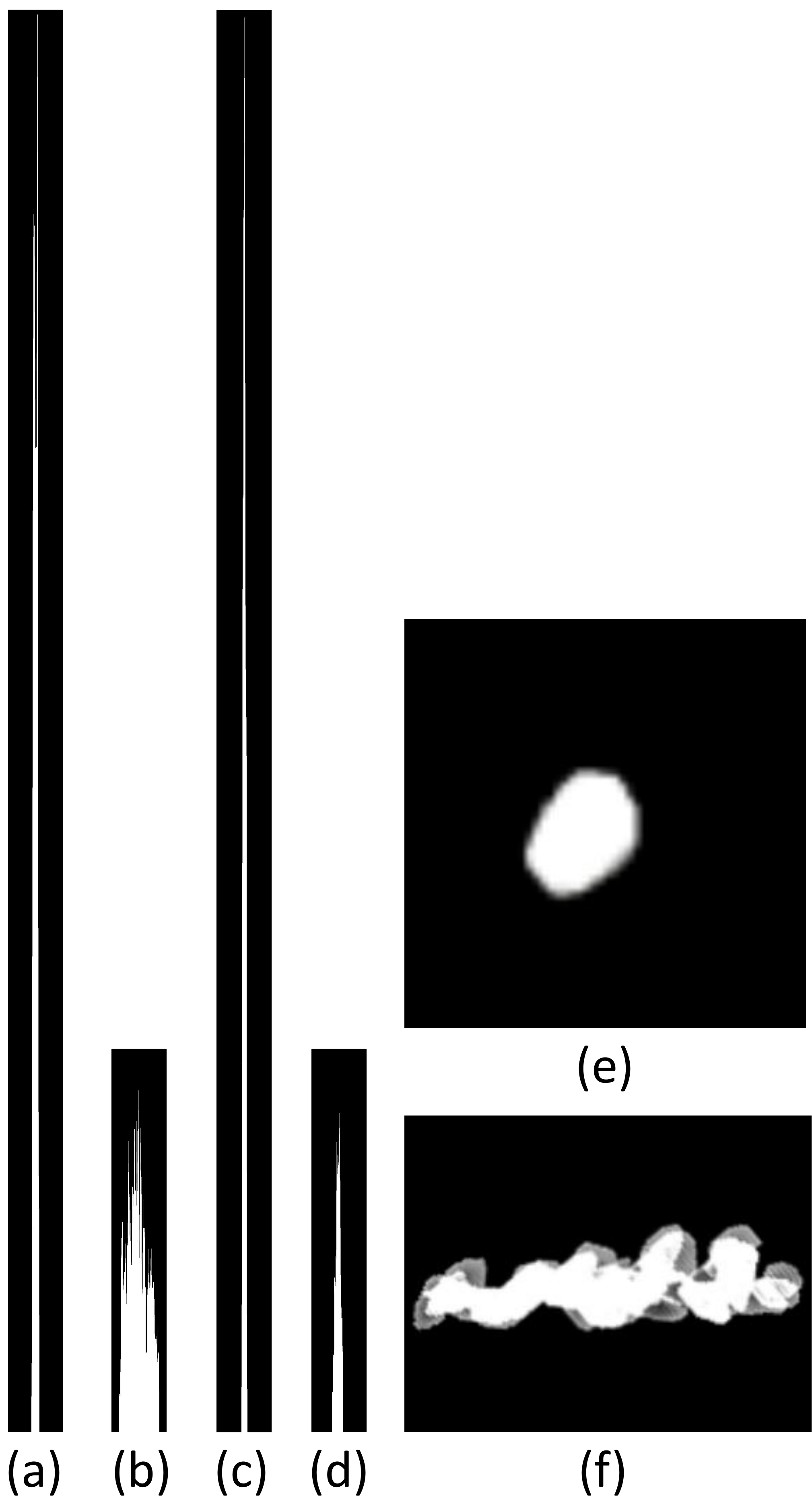}}
\caption{2D visualization of two foldovers in the X, Y, and Z directions. (a) (c) (e) are the foldovers of a nearly stationary sperm in the three directions, therefore, the foldovers of this sperm in the X and Y directions are significantly higher than the others. (b) (d) (f) are the foldovers of the swimming sperm. }
\label{fig:different sperm foldoverXYZ}
\end{figure}

So, the grade of \figurename~\ref{fig:different sperm foldover} (a) is D (immotility), and the grade of \figurename~\ref{fig:different sperm foldover} (b) is A (rapid progressive motility). In X direction, the foldover contains the range of moving direction, which is the length of the foldover along the X direction $\digamma_{\left( {i,g} \right)}^R\left( \rm{X}  \right)$. The total number of frames ${\Gamma _{\left( {i,g} \right)}}\left( {{O_{\left( {i,g} \right)}}} \right)$ that make up the foldover $\digamma_{\left( {i,g} \right)}$ is the time information, which is the movement time of sperm, and by ${\Gamma _{\left( {i,g} \right)}}\left( {{O_{\left( {i,g} \right)}}} \right)$ we calculate the frame rate of $\digamma_{\left( {i,g} \right)}$ in the X direction. We define the frame rate as $v_{\left( {i,g} \right)}^{\left( {{\rm{FPS}},{\rm{X}}} \right)}$ in Eq.~(\ref{con: eq-10}) , and 2D visualization of two foldovers in the X direction are shown in (a) and (b) of \figurename~\ref{fig:different sperm foldoverXYZ}.

\begin{equation}
\label{con: eq-10}
v_{\left( {i,g} \right)}^{\left( {{\rm{FPS}},{\rm{X}}} \right)} = \frac{{\digamma_{\left( {i,g} \right)}^R\left( {\rm{X}} \right)}}{{{\Gamma _{\left( {i,g} \right)}}\left( {{O_{\left( {i,g} \right)}}} \right)}}
\end{equation}

In Y direction, the foldover contains the range of the orthogonal direction of moving direction, which is the length of the foldover along the Y direction $\digamma_{\left( {i,g} \right)}^R\left( \rm{Y}  \right)$. Similar to X direction, we calculate the frame rate of $\digamma_{\left( {i,g} \right)}$ in the Y direction by ${\Gamma _{\left( {i,g} \right)}}\left( {{O_{\left( {i,g} \right)}}} \right)$, we define the frame rate as $v_{\left( {i,g} \right)}^{\left( {{\rm{FPS}},{\rm{Y}}} \right)}$ in Eq.~(\ref{con: eq-11}) , and 2D visualization of two foldovers in the Y direction are shown in (c) and (d) of \figurename~\ref{fig:different sperm foldoverXYZ}.

\begin{equation}
\label{con: eq-11}
v_{\left( {i,g} \right)}^{\left( {{\rm{FPS}},{\rm{Y}}} \right)} = \frac{{\digamma_{\left( {i,g} \right)}^R\left( {\rm{Y}} \right)}}{{{\Gamma _{\left( {i,g} \right)}}\left( {{O_{\left( {i,g} \right)}}} \right)}}
\end{equation}

In Z direction, the foldover contains trajectory, shape, and brightness information. By the trajectory of the foldover we calculate the motion distance, the motion displacement and the average path length. Furthermore, we calculate the motion distance and the motion displacement by ${\phi _{\left( i \right)}} = \left\{ {{S_{\left( {i,1} \right)}},{S_{\left( {i,2} \right)}},...,{S_{\left( {i,g} \right)}},...,{S_{\left( {i,\tau} \right)}}} \right\}$ ( ${S_{\left( {i,g} \right)}} = \left\{ {{I_{\left( {i,j,g} \right)}},} \right.$ $\left. {{I_{\left( {i,j + 1,g} \right)}},...} \right\}$ ), and by fitting ${\phi _{\left( i \right)}} = \left\{ {{S_{\left( {i,1} \right)}},{S_{\left( {i,2} \right)}},...,{S_{\left( {i,g} \right)}},...,{S_{\left( {i,\tau} \right)}}} \right\}$ to the third power, an equation can be calculated based on the motion path, then the average path length of sperm is calculated by combining this equation. We define the motion distance as ${A_{\left( {i,g} \right)}}$, the motion displacement as ${B_{\left( {i,g} \right)}}$, the fitted equation as $\varrho \left( {{I_{\left( {i,j,g} \right)}}} \right)$ and the average path length as ${M_{\left( {i,g} \right)}}$, the formula of ${A_{\left( {i,g} \right)}}$, ${B_{\left( {i,g} \right)}}$ and ${M_{\left( {i,g} \right)}}$ are expressed by Eq.~(\ref{con: eq-12}) , Eq.~(\ref{con: eq-13}) and Eq.~(\ref{con: eq-14}) .

\begin{equation}
\label{con: eq-12}
{A_{\left( {i,g} \right)}} = \sum\limits_j^{{\Gamma _{\left( {i,g} \right)}}\left( {{O_{\left( {i,g} \right)}}} \right) - 1} {\left[ {{I_{\left( {i,j + 1,g} \right)}} - {I_{\left( {i,j + 1,g} \right)}}} \right]} 
\end{equation}

In Eq.~(\ref{con: eq-12}) , we add up all the barycentric coordinates ${S_{\left( {i,g} \right)}} = \left\{ {{I_{\left( {i,j,g} \right)}},} \right.$ $\left. {{I_{\left( {i,j + 1,g} \right)}},...} \right\}$ contained in the foldover as the distance of motion ${A_{\left( {i,g} \right)}}$.

\begin{equation}
\label{con: eq-13}
{B_{\left( {i,g} \right)}} = {I_{\left( {i,{\Gamma _{\left( {i,g} \right)}}\left( {{O_{\left( {i,g} \right)}}} \right),g} \right)}} - {I_{\left( {i,j,g} \right)}}
\end{equation}

In Eq.~(\ref{con: eq-13}) , we calculate the distance between the first position ${I_{\left( {i,j,g} \right)}}$ and the last position ${I_{\left( {i,{\Gamma _{\left( {i,g} \right)}}\left( {{O_{\left( {i,g} \right)}}} \right),g} \right)}}$ of the foldover as the motion displacement ${B_{\left( {i,g} \right)}}$.

\begin{equation}
\label{con: eq-14}
{M_{\left( {i,g} \right)}} = \sum\limits_j^{{\Gamma _{\left( {i,g} \right)}}\left( {{O_{\left( {i,g} \right)}}} \right){\rm{ - 1}}} {\left[ {\varrho \left( {{I_{\left( {i,j{\rm{ + 1}},g} \right)}}} \right){\rm{ - }}\varrho \left( {{I_{\left( {i,j,g} \right)}}} \right)} \right]}
\end{equation}

In Eq.~(\ref{con: eq-14}) , by fitting the equation ${\varrho \left( {{I_{\left( {i,j,g} \right)}}} \right)}$, we can calculate the new coordinates corresponding to ${S_{\left( {i,g} \right)}} = \left\{ {{I_{\left( {i,j,g} \right)}},} \right.$ $\left. {{I_{\left( {i,j + 1,g} \right)}},...} \right\}$, and add the distance between these new coordinates we can obtain the average path length ${M_{\left( {i,g} \right)}}$.

According to the motion distance ${A_{\left( {i,g} \right)}}$, motion displacement ${B_{\left( {i,g} \right)}}$ and average path length ${M_{\left( {i,g} \right)}}$, we can further calculate curvilinear velocity ($\text{VCL}$) , straight line velocity ($\text{VSL}$) and average path velocity ($\text{VAP}$) in Eq.~(\ref{con: eq-15}) .

\begin{equation}
\label{con: eq-15}
\begin{split}
  & v_{\left( i,g \right)}^{\left( \text{VCL} \right)}\text{=}\frac{{{A}_{\left( i,g \right)}}}{{{\Gamma }_{\left( i,g \right)}}\left( {{O}_{\left( i,g \right)}} \right)} \\ 
 & v_{\left( i,g \right)}^{\left( \text{VSL} \right)}\text{=}\frac{{{B}_{\left( i,g \right)}}}{{{\Gamma }_{\left( i,g \right)}}\left( {{O}_{\left( i,g \right)}} \right)} \\ 
 & v_{\left( i,g \right)}^{\left( \text{VAP} \right)}\text{=}\frac{{{M}_{\left( i,g \right)}}}{{{\Gamma }_{\left( i,g \right)}}\left( {{O}_{\left( i,g \right)}} \right)} \\ 
\end{split}
\end{equation}

We define the $\text{VCL}$ as $v_{\left( {i,g} \right)}^{\left( {{\rm{VCL}}} \right)}$, and we obtain $\text{VCL}$ based on ${{A_{\left( {i,g} \right)}}}$ and ${{\Gamma _{\left( {i,g} \right)}}\left( {{O_{\left( {i,g} \right)}}} \right)}$. The $\text{VSL}$ is defined as $v_{\left( {i,g} \right)}^{\left( {{\rm{VSL}}} \right)}$, we calculate the $\text{VSL}$ based on ${{B_{\left( {i,g} \right)}}}$ and ${{\Gamma _{\left( {i,g} \right)}}\left( {{O_{\left( {i,g} \right)}}} \right)}$. The $\text{VAP}$ is defined as $v_{\left( {i,g} \right)}^{\left( {{\rm{VAP}}} \right)}$, we calculate the $\text{VAP}$ based on ${{M_{\left( {i,g} \right)}}}$ and ${{\Gamma _{\left( {i,g} \right)}}\left( {{O_{\left( {i,g} \right)}}} \right)}$.

Furthermore, using $v_{\left( {i,g} \right)}^{\left( {{\rm{VCL}}} \right)}$, $v_{\left( {i,g} \right)}^{\left( {{\rm{VSL}}} \right)}$ and $v_{\left( {i,g} \right)}^{\left( {{\rm{VAP}}} \right)}$, we can calculate linearity ($\text{LIN}$) , Straightness ($\text{STR}$) and Wobble ($\text{WOB}$) in Eq.~(\ref{con: eq-16}) .

\begin{equation}
\label{con: eq-16}
\begin{split}
  & \text{LI}{{\text{N}}_{\left( i,g \right)}}=\frac{v_{\left( i,g \right)}^{\left( \text{VSL} \right)}}{v_{\left( i,g \right)}^{\left( \text{VCL} \right)}} \\ 
 & \text{ST}{{\text{R}}_{\left( i,g \right)}}=\frac{v_{\left( i,g \right)}^{\left( \text{VSL} \right)}}{v_{\left( i,g \right)}^{\left( \text{VAP} \right)}} \\ 
 & \text{WO}{{\text{B}}_{\left( i,g \right)}}=\frac{v_{\left( i,g \right)}^{\left( \text{VAP} \right)}}{v_{\left( i,g \right)}^{\left( \text{VCL} \right)}} \\ 
\end{split}
\end{equation}

$\text{LI}{{\text{N}}_{\left( i,g \right)}}$ is the ratio of ${v_{\left( {i,g} \right)}^{\left( {{\rm{VSL}}} \right)}}$ to ${v_{\left( {i,g} \right)}^{\left( {{\rm{VCL}}} \right)}}$, $\text{ST}{{\text{R}}_{\left( i,g \right)}}$ is the ratio of ${v_{\left( {i,g} \right)}^{\left( {{\rm{VSL}}} \right)}}$ to ${v_{\left( {i,g} \right)}^{\left( {{\rm{VAP}}} \right)}}$, and $\text{WO}{{\text{B}}_{\left( i,g \right)}}$ is the ratio of ${v_{\left( {i,g} \right)}^{\left( {{\rm{VAP}}} \right)}}$ to ${v_{\left( {i,g} \right)}^{\left( {{\rm{VCL}}} \right)}}$.

Regarding the shape information, foldovers can detect the deformation of sperm during the movement. The brightness information mainly includes the pixel accumulation process, the higher brightness area indicates that the sperm stay in this area for the longer time. 2D visualization of two foldovers in the Z direction are shown in (e) and (f) of \figurename~\ref{fig:different sperm foldoverXYZ}

Although $U\left( {\digamma_{\left( {i,g} \right)}^{\left( {R,X} \right)}} \right)$, $U\left( {\digamma_{\left( {i,g} \right)}^{\left( {R,Y} \right)}} \right)$ and $U\left( {\digamma_{\left( {i,g} \right)}^{\left( {R,Z} \right)}} \right)$ include the information of foldovers, they are three matrices of an object (such as a sperm) with a lot of redundant information. Therefore, we make statistics on all the information of $U\left( {\digamma_{\left( {i,g} \right)}^{\left( {R,X} \right)}} \right)$, $U\left( {\digamma_{\left( {i,g} \right)}^{\left( {R,Y} \right)}} \right)$ and $U\left( {\digamma_{\left( {i,g} \right)}^{\left( {R,Z} \right)}} \right)$ to optimize them. Especially, we apply convolutional operations to achieve the optimization, where we define the process of convolution optimization as $H^\Theta$ ($\Theta$ = X, Y and Z), $H_{\left( {{\rm{i}},{\rm{g}},{\rm{k}}} \right)}^\Theta$ is the $k$-th pixel of the $g$-th foldover in the $i$-th video, and $H$ is defined in Eq.~(\ref{con: eq-17}) .

\begin{equation}
\begin{aligned}
\label{con: eq-17}
H_{\left( \text{i},\text{g},\text{k} \right)}^{\Theta }=U*G=\sum\limits_{e}{p\left[ {{U}_{\left( i,g,k \right)}}\left( \digamma _{\left( i,g \right)}^{\left( R,\Theta  \right)} \right) \right]}p\left( {{G}_{\left( k-e \right)}} \right)
\end{aligned}
\end{equation}

In Eq.~(\ref{con: eq-17}) , $G$ is the convolution kernel, and $e$ is the dimension of the $G$. Here, because we cannot consolidate all the useful information and get rid of all the redundant information by just once convolution, we need to do multiple convolutions.

Furthermore, $v_{\left( {i,g} \right)}^{\left( {{\rm{FPS}},{\rm{X}}} \right)}$, $v_{\left( {i,g} \right)}^{\left( {{\rm{FPS}},{\rm{Y}}} \right)}$, ${A_{\left( {i,g} \right)}}$, ${B_{\left( {i,g} \right)}}$, ${M_{\left( {i,g} \right)}}$, $v_{\left( {i,g} \right)}^{\left( {{\rm{VCL}}} \right)}$, $v_{\left( {i,g} \right)}^{\left( {{\rm{VSL}}} \right)}$, $v_{\left( {i,g} \right)}^{\left( {{\rm{VAP}}} \right)}$, $\text{LI}{{\text{N}}_{\left( i,g \right)}}$, $\text{ST}{{\text{R}}_{\left( i,g \right)}}$, $\text{WO}{{\text{B}}_{\left( i,g \right)}}$ and $H_{\left( {{\rm{i}},{\rm{g}},{\rm{k}}} \right)}^\Theta$ are joined together to form three foldover feature vectors, where $v_{\left( {i,g} \right)}^{\left( {{\rm{FPS}},{\rm{X}}} \right)}$ and $H_{\left( {{\rm{i}},{\rm{g}},{\rm{k}}} \right)}^{\rm{X}}$ are concatenated to form the foldover feature $F_{\left( i,g \right)}^{\rm{X}}$ of the X direction; $v_{\left( {i,g} \right)}^{\left( {{\rm{FPS}},{\rm{Y}}} \right)}$ and $H_{\left( {{\rm{i}},{\rm{g}},{\rm{k}}} \right)}^{\rm{Y}}$ are concatenated to form the foldover feature $F_{\left( i,g \right)}^{\rm{Y}}$ of the Y direction; ${A_{\left( {i,g} \right)}}$, ${B_{\left( {i,g} \right)}}$, ${M_{\left( {i,g} \right)}}$, $v_{\left( {i,g} \right)}^{\left( {{\rm{VCL}}} \right)}$, $v_{\left( {i,g} \right)}^{\left( {{\rm{VSL}}} \right)}$, $v_{\left( {i,g} \right)}^{\left( {{\rm{VAP}}} \right)}$, $\text{LI}{{\text{N}}_{\left( i,g \right)}}$, $\text{ST}{{\text{R}}_{\left( i,g \right)}}$, $\text{WO}{{\text{B}}_{\left( i,g \right)}}$ and $H_{\left( {{\rm{i}},{\rm{g}},{\rm{k}}} \right)}^{\rm{Z}}$ are concatenated to form the foldover feature $F_{\left( i,g \right)}^{\rm{Z}}$ of the Z direction. The algorithm of the foldover features are shown in Algorithm~\ref{alg:Generation of H}.

\begin{algorithm}[H]  
\caption{Generation of $H_{\left( {{\rm{i}},{\rm{g}},{\rm{k}}} \right)}^\Theta$}  
\label{alg:Generation of H}  
\hspace*{0.02in}{\bf Input:}
Videos $\chi$\\preprocessed video $X_i$\\
\hspace*{0.02in}{\bf Output:} 
 $H_{\left( {{\rm{i}},{\rm{g}},{\rm{k}}} \right)}^\Theta$, $\Theta$ = X, Y and Z 
\begin{algorithmic}[1] 
\STATE {video decomposition: \\${X_i} \!=\! \left\{ {{x_{\left( {i,1} \right)}},{x_{\left( {i,2} \right)}},...,{x_{\left( {i,j} \right)}},...,{x_{\left( {i,m} \right)}}} \right\}$}
\STATE {image segmentation: $x_{\left( {i,j} \right)}^{seg} \Leftarrow x_{\left( {i,j} \right)}$ \\$p\left( {{x}^{\rm{seg}}}_{\left( i,j,k \right)} \right)=\left\{ \begin{matrix}
   0 & p\left( {{x}_{\left( i,j,k \right)}} \right)\le T\left( {{x}_{\left( i,j \right)}} \right)  \\
   1 & otherwise  \\
\end{matrix} \right.$}
\STATE {barycenter coordinates extraction: \\${{\psi }_{\left( i \right)}}=\left\{ {{C}_{\left( i,1 \right)}},{{C}_{\left( i,2 \right)}},...,{{C}_{\left( i,j \right)}},...,{{C}_{\left( i,m \right)}} \right\}$}
\STATE {target matching: \\${{d}_{\left( i,j,l \right)}}=\sqrt{{{\left[ c\left( {{s}_{\left( i,j+1,l \right)}} \right)-c\left( {{s}_{\left( i,j,l \right)}} \right) \right]}^{2}}}$}
\STATE {construction of the foldover: \\$p\left[ {{\digamma_{\left( {i,g} \right)}}\left( {{x_{\left( {i,j,k} \right)}}} \right)} \right]{\rm{ = }}\sum\limits_j^{{\Gamma _{\left( {i,g} \right)}}\left( {{O_{\left( {i,g} \right)}}} \right)} {p\left[ {{o_{\left( {i,j,g} \right)}}\left( {{x_{\left( {i,j,k} \right)}}} \right)} \right]}$}
\STATE {rotate the foldover : $\digamma_{\left( {i,g} \right)}^R \Leftarrow {\digamma _{\left( {i,g} \right)}}$}
\STATE {foldover processing: $\digamma_{\left( {i,g} \right)}^{\left( {R,\Theta} \right)} \Leftarrow \digamma_{\left( {i,g} \right)}^R$ \\$U\left( {\digamma_{\left( {i,g} \right)}^{\left( {R,\Theta } \right)}} \right) = \sum\limits_{u = 1}^{\frac{{\digamma_{\left( {i,g} \right)}^{\left( R \right)}\left( \Theta  \right)}}{{{\nu _\Theta }}}} {\digamma_{\left( {i,g,u} \right)}^{\left( {R,\Theta } \right)}}$ }
\STATE {the optimization of $U\left( {\digamma_{\left( {i,g} \right)}^{\left( {R,\Theta } \right)}} \right)$: \\${H_{\left( {{\rm{i}},{\rm{g}},{\rm{k}}} \right)}^\Theta} = U * G$}
\STATE {the generation of foldover features: $F_{\left( i,g \right)}^{\text{X}}$, $F_{\left( i,g \right)}^{\text{Y}}$, $F_{\left( i,g \right)}^{\text{Z}}$}
\end{algorithmic}  
\end{algorithm} 

Finally, We obtain the foldover feature vectors, $F_{\left( i,g \right)}^{\text{X}}$, $F_{\left( i,g \right)}^{\text{Y}}$ and $F_{\left( i,g \right)}^{\text{Z}}$. According to the foldover features, we can solve the following difficulties we encounter in the microscopic videos: (1) Multi-object recognition, (2) Similar object recognition, (3) Tiny object recognition, (4) Impurity interference and (5) Little feature information.

\section{Experimental Results and Analysis}
\label{s:experiments}

In this section, experimental results and analysis are discussed, including \ref{s:Experimental Setting} experimental setting, \ref{s:Experimental Results} experimental results. 

\subsection{Experimental Setting}
\label{s:Experimental Setting}
\subsubsection{Experimental Data}
In this paper, a practical microscopic video set $\chi {\rm{=}}\left\{ {{X_1},{X_{\rm{2}}}, \ldots ,{X_i},...,{X_{59}}} \right\}$ with 59 semen videos is applied to test our method. The format of the videos is grey-scale mp4, the size of each frame is 698$\times$528$\times$3 pixels and the frame rate is 30 frames per second ( FPS ). There are 1,374 sperms in set $\chi$. For all the sperms, ground truth (GT) images are prepared manually by four biomedical engineers and two medical doctors, where the sperms are labeled as foreground object with 1 (white) and other regions are labeled as background with 0 (black). We mark the number of each sperm in the video and propose the following strategy for sperm numbering:
\begin{itemize}
\item {\bfseries Case-I: } All the sperms in the video ( moving or stationary ) are numbered, the numbers increased from 1, and  each sperm is numbered horizontally from the top of the visual field.
\item {\bfseries Case-II: } If there is a sperm swimming out of the visual field, we stipulate that the motion of this sperm is over.
\item {\bfseries Case-III: } If there is a sperm swimming into the visual field, we assume we have a new sample and give it a new number.
\item {\bfseries Case-IV: } A malformed sperm is considered a sample, for example a sperm with two heads or two tails.
\end{itemize} 

\begin{figure}[H]
\centering
\includegraphics[scale=0.4]{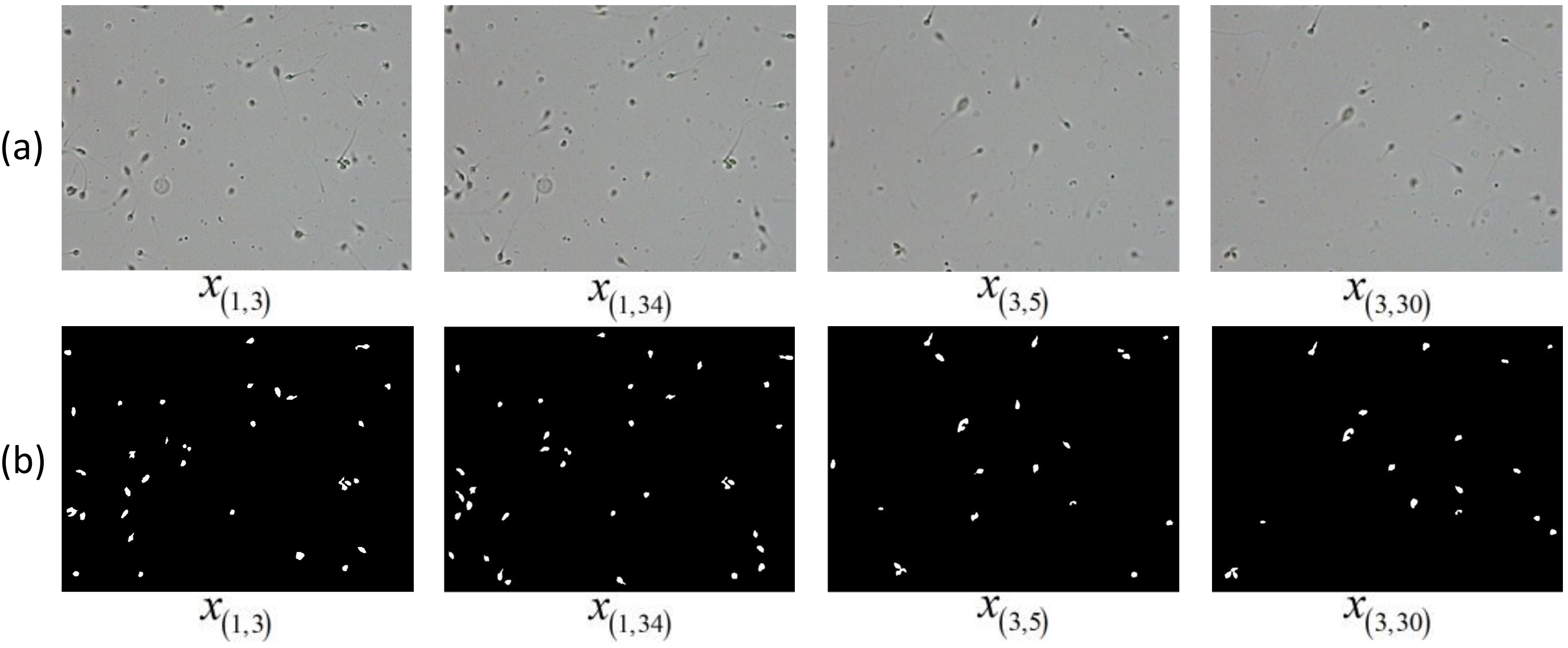}
\caption{An example of frames and their GT images in a semen microscopic video. (a) shows the frames and (b) shows the GT images.}
\label{fig:frame_and_gray}
\end{figure}

Furthermore, based on the diagnosis of the medical doctors, all sperms are grouped into three classes, including poor motion state, good motion state and excellent motion state. There are 950 samples of poor motion state, 262 samples of good motion state and 162 samples of excellent motion state. In addition, the number of samples in the training set is equal to that in the testing set, 687 samples are used for the training set and 687 samples are used for the testing set. In the training set, the sample number of poor motion state is 462, the sample number of good motion state is 138, and the sample number of excellent motion state is 87. In the testing set, the sample number of poor motion state is 488, the sample number of good motion state is 124, and the sample number of excellent motion state is 75. An example of the video frames and their GT images is shown in \figurename~\ref{fig:frame_and_gray}.

\subsubsection{Evaluation Index}
We use classifiers to evaluate foldover features with a three-class classification task of sperms, and the classification evaluation indicators are shown in TABLE 1~\cite{Lamb-WHO-2000}. Specifically, four classifiers are tested in this paper, including Artificial Neural Networks~\cite{Judith-ANN-2001} (ANNs), Random Forests~\cite{Ho-RF-1995} (RFs) and Support Vector Machines~\cite{Saunders-SVM-2002} (linear-SVM and RBF-SVM). Because there are more motionless, slow-swimming sperms and fewer fast-swimming sperms in the videos, we calculate multiple indexes to evaluate the proposed foldover features. Firstly, we calculate the confusion matrix of all classification results. Then, based on the confusion matrices, we can further calculate the accuracy, precision, recall, specificity and F1-measure as shown in Table~\ref{tab: Indicators1}. 

\begin{table}[H]
\centering
\normalsize 
\caption{The evaluation of confusion matrix}
\begin{tabular}{|c|c|}
\hline
\textbf{Indicators} & \textbf{Formulas} \\ \hline
Accuracy            & ${\left( \text{TP}+\text{TN} \right)}/{\left( \text{P}+\text{N} \right)}\;$ \\ \hline
Precision           & ${\text{TP}}/{\left( \text{TP}+\text{FP} \right)}\;$                 \\ \hline
Recall              & ${\text{TP}}/{\left( \text{TP}+\text{FN} \right)}\;$                 \\ \hline
Specificity       & ${\text{TN}}/{\left( \text{TN}+\text{FP} \right)}\;$                 \\ \hline
F1-measure        & ${\text{2TP}}/{\left( \text{2TP}+\text{FP}+\text{FN} \right)}\;$                 \\ \hline
\end{tabular}
\label{tab: Indicators1}
\end{table}

The negative number of the actual sample is N=TN+FP, the number of positive is P=FN+TP, and the total sample size is C=N+P, where TP is True Positive, TN is True Negative, FP is False Positive and FN is False Negative. Recall (also known as sensitivity) can measure the reliability of the model's prediction with the positive sample, a higher recall means that an algorithm returns more relevant results. Precision can measure the accuracy of the model in predicting positive samples, a higher precision means that an algorithm returns substantially more relevant results than irrelevant ones. Specificity (also called the TN rate) measures the proportion of actual negatives that are correctly identified as such. F1-measure is a measure of an accuracy of a test, considering both the precision and the recall of the test to compute the score.

Thirdly, because our experiment is used for three categories, the precision has three values, and each class has its corresponding precision, we define the three values of precision as $\text{Precision}1$, $\text{Precision}2$ and $\text{Precision}3$. In the same way, there are also three values for recall defined as $\text{Recall}1$, $\text{Recall}2$, $\text{Recall}3$. Based on the confusion matrices, we can calculate the macro precision, the macro recall and the macro F1-measure as shown in Table~\ref{tab: Indicators2}.

\begin{table}[H]
\centering
\normalsize 
\caption{The evaluation of three categories}
\begin{tabular}{|c|c|}
\hline
\textbf{Indicators} & \textbf{Formulas} \\ \hline
Macro\_P             & ${1}/{3}\;\left( \text{Precision}1+\text{Precision}2+\text{Precision}3 \right)$               \\ \hline
Macro\_R             & ${1}/{3}\;\left( \text{Recall}1+\text{Recall}2+\text{Recall}3 \right)$                 \\ \hline
Macro\_F1           & ${\left( 2\times \text{Macro}\_\text{P}\times \text{Macro}\_\text{R} \right)}/{\left( \text{Macro}\_\text{P+Macro}\_\text{R} \right)}\;$                 \\ \hline
Variance            & ${1}/{3}\;\left[ {{\left( \text{Recall}1-\text{Macro}\_\text{R} \right)}^{2}}+{{\left( \text{Recall2}-\text{Macro}\_\text{R} \right)}^{2}}+{{\left( \text{Recall3}-\text{Macro}\_\text{R} \right)}^{2}} \right]$                 \\ \hline
\end{tabular}
\label{tab: Indicators2}
\end{table}

Because our experiment is a triage experiment, therefore, when we calculate $\text{Macro}\_\text{P}$, we need to calculate the mean of $\text{Precision}1$, $\text{Precision}2$ and $\text{Precision}3$, and the calculation of $\text{Macro}\_\text{R}$ is the same. Finally, based on the accuracy of each category, we calculate the varianceas shown in Table~\ref{tab: Indicators2}.

\subsection{Experimental Results}
\label{s:Experimental Results}
\subsubsection{Evaluation for Foldover Features}
Artificial Neural Networks~\cite{Judith-ANN-2001} (ANNs), Random Forests~\cite{Ho-RF-1995} (RFs) and Support Vector Machine~\cite{Saunders-SVM-2002} (linear-SVM and RBF-SVM) are used to test the effectiveness of the foldover features. Specifically, the parameters of the ANNs are set as follows: The number of network layers is 2, the number of hidden nodes is 10, and the activation function is log-sigmoid; The parameter of the RFs is set as follows: The number of decision tree is 200; The parameters of the Support Vector Machine are set as follows: Kernel function of linear-SVM is linear kernel, kernel function of RBF-SVM is radial basis function.

\begin{figure*}[htbp]
\centering
\includegraphics[scale=0.19]{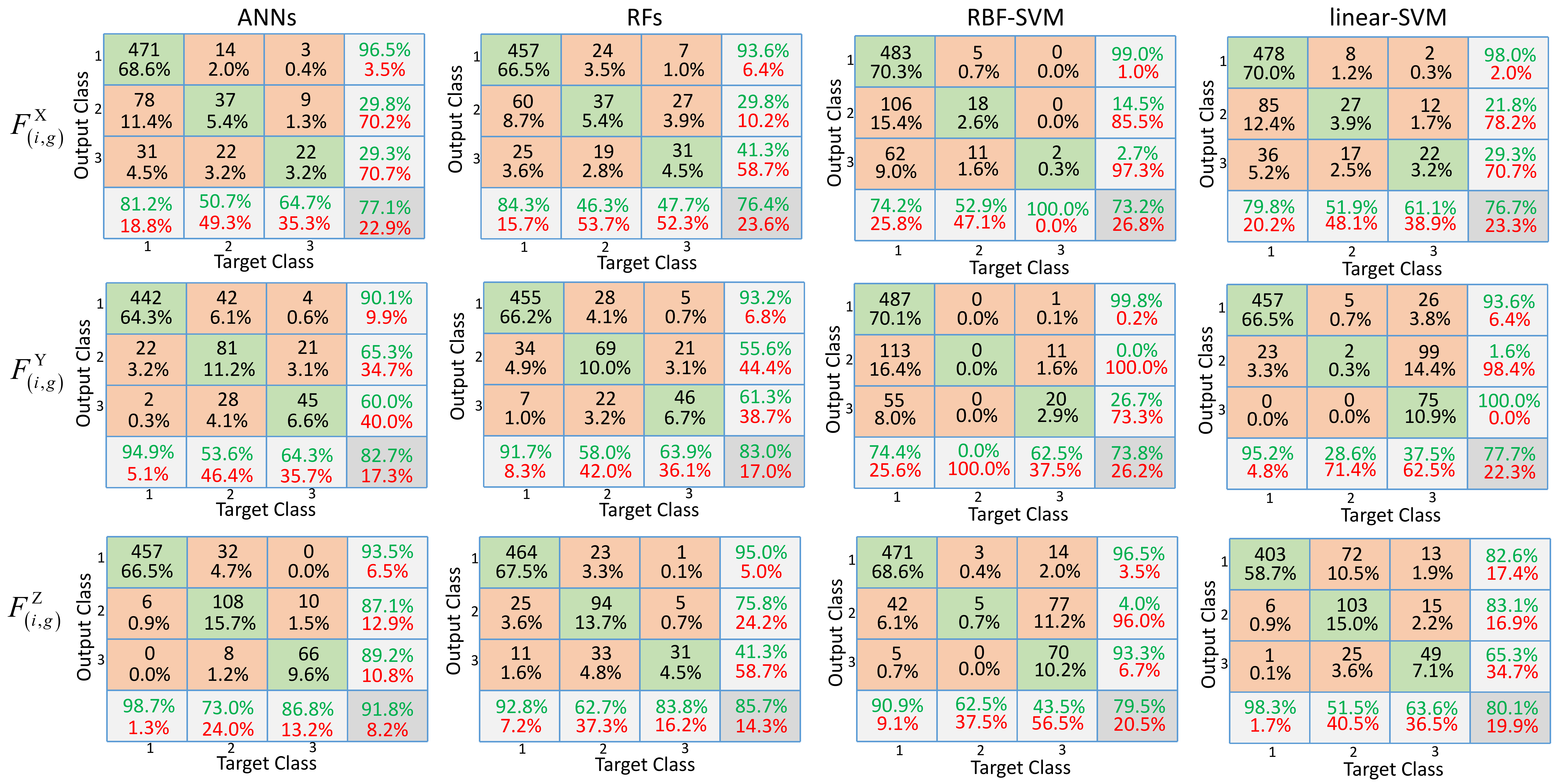}
\caption{The confusion matrices of $F_{\left( i,g \right)}^{\text{X}}$, $F_{\left( i,g \right)}^{\text{Y}}$ and $F_{\left( i,g \right)}^{\text{Z}}$. Rows represent the foldover features, and columns represent the classifiers.}
\label{fig:XYZ_foldover_confusion_matrix}
\end{figure*}

The foldover features, $F_{\left( i,g \right)}^{\text{X}}$, $F_{\left( i,g \right)}^{\text{Y}}$ and $F_{\left( i,g \right)}^{\text{Z}}$ are classified by ANNs, RFs, linear-SVM and RBF-SVM, and the confusion matrices of classification results are shown in \figurename~\ref{fig:XYZ_foldover_confusion_matrix}.

$F_{\left( i,g \right)}^{\text{Z}}$ obtains the best results in four classifiers, especially in ANNs, the accuracy is 91.8\%, and the classification accuracy of each category is also excellent, 93.5\%, 87.1\% and 89.2\%, respectively.

\subsubsection{Comparison with Static Features}
\begin{figure}[H]
\centering
\includegraphics[scale=0.3]{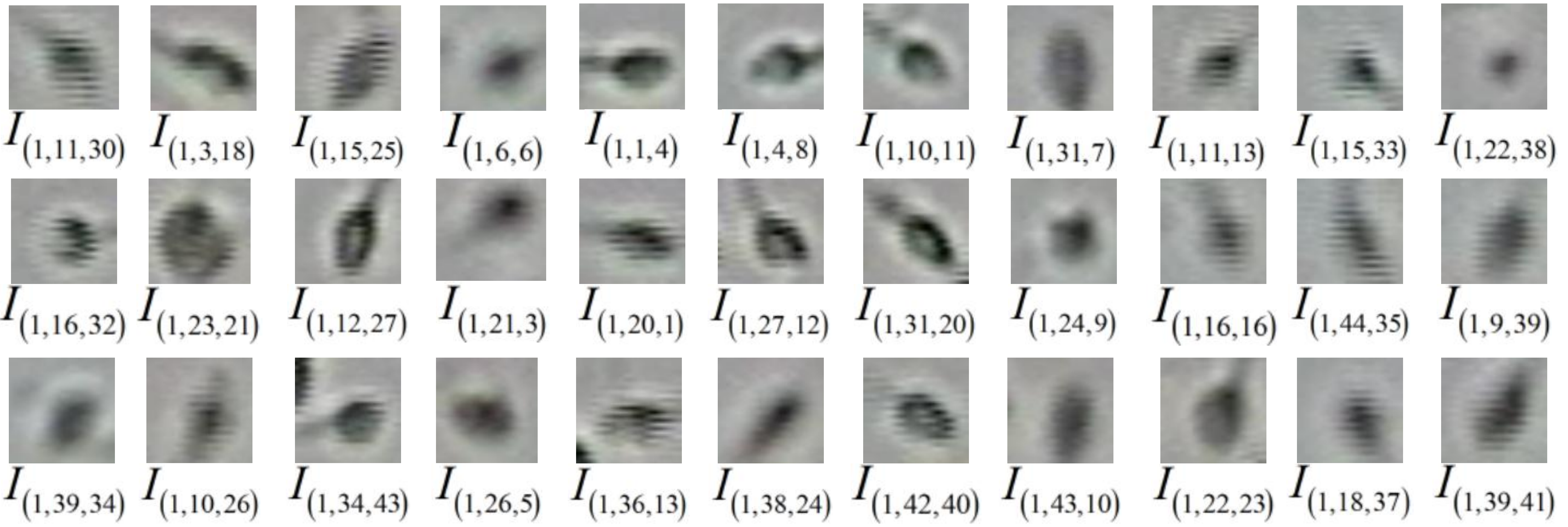}
\caption{An example of detected sperms.}
\label{fig:detected_sperms}
\end{figure}

Firstly, according to the ${\phi _{\left( i \right)}} = \left\{ {{S_{\left( {i,1} \right)}},{S_{\left( {i,2} \right)}},...,{S_{\left( {i,g} \right)}},...,{S_{\left( {i,\tau} \right)}}} \right\}$, each sperm is detected to a size of 26 by 26 pixels in the corresponding frame, the pixel size of 26 by 26 is an ideal size after repeated experiments to ensure which is the only sperm we want in the detected image, and an example of some detected sperms is shown in \figurename~\ref{fig:detected_sperms}.

\begin{figure*}[htbp]
\centering
\includegraphics[scale=0.22]{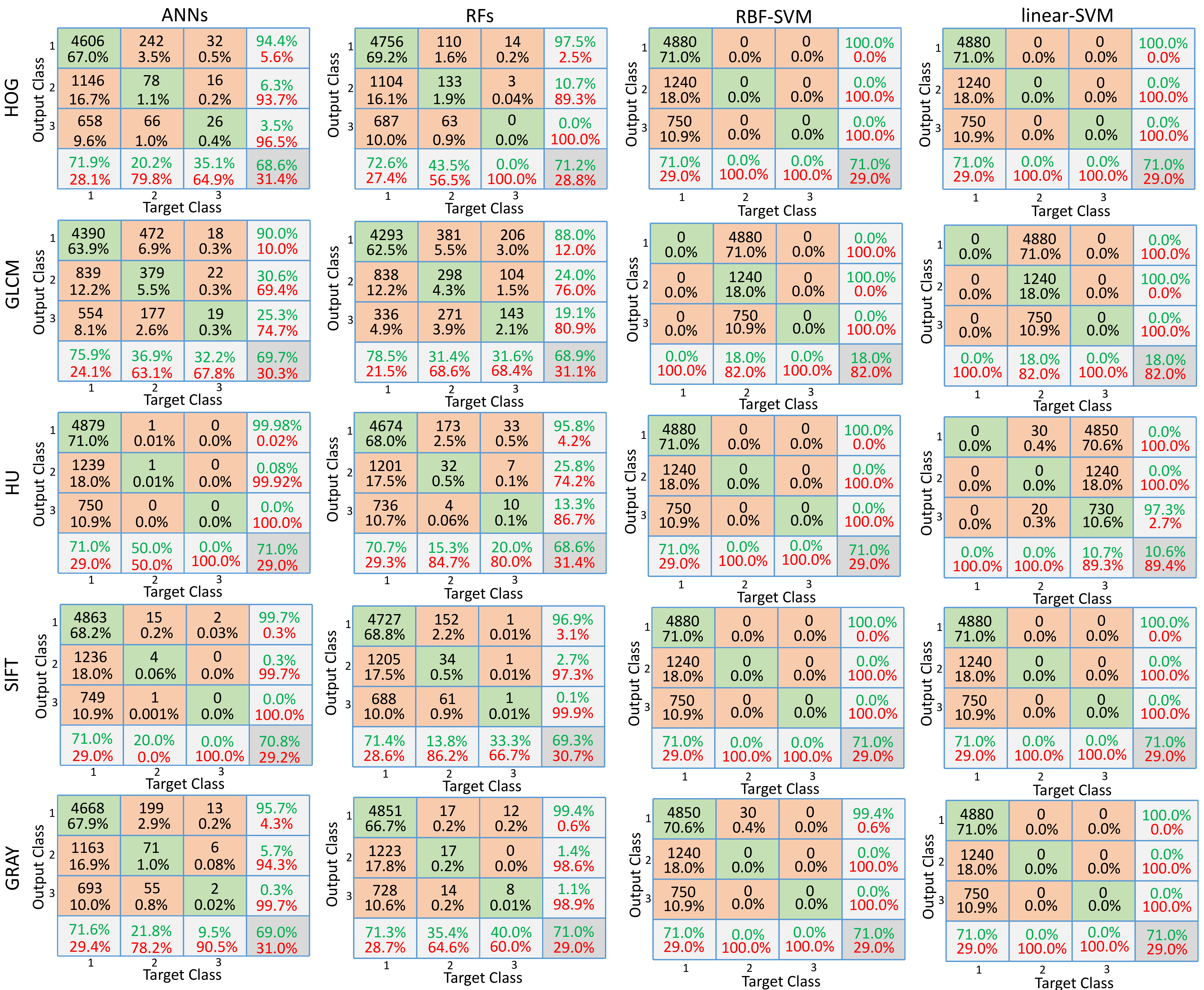}
\caption{Confusion matrices of the static features. Rows represent feature types, and columns represent classifier types.}
\label{fig:static_feature_confusion_matrix}
\end{figure*}

Secondly, we extract the static features of sperms after detection, including Histogram of Oriented Gradient~\cite{NavneetDalal-2005-HOG} (HOG), Grey-Level Co-occurrence Matrix~\cite{Seongjin-2011-GGO} (GLCM), the geometric invariant moment proposed by Hu~\cite{Hu-1962-Visual}, Scale-Invariant Feature Transform~\cite{Mortensen-2005-SIFT} (SIFT) and gray histogram~\cite{Brunelli-2002-OTS}. All static features are extracted from detected sperm images, but the movement of a sperm exists in multiple frames, therefore, we adopt the method of multiple extraction, and randomly select one sperm image from all the images of this sperm at a time to extract the static features. Thirdly, the number of times to extract the static feature is ten, obviously, the number of times to classify the static feature is ten. We use Artificial Neural Networks~\cite{Judith-ANN-2001} (ANNs), Random Forests~\cite{Ho-RF-1995} (RFs) and Support Vector Machine~\cite{Saunders-SVM-2002} (linear-SVM and RBF-SVM)) classifiers to classify static features and construct the total confusion matrices of ten experiments to represent the classification results. The classification results of static features in four classifiers are shown in in \figurename~\ref{fig:static_feature_confusion_matrix}.

According to the confusion matrices of static features in \figurename~\ref{fig:static_feature_confusion_matrix}, we calculate the evaluations, and the comparison evaluations between static features and foldover features are shown in TABLE\ref{tab: static features and foldover features}.

\begin{table*}[htbp]
\tiny
\caption{Evaluation of static features with four classifiers. The first column shows the types of static features, the second column shows the types of classifiers, the third to the last columns show the calculated evaluations. We use the first three letters of each evaluation to indicate the evaluation metric, such as Acc is accuracy, Pre is precision, Mac\_P is Macro\_P, Rec is recall, Mac\_R is Macro\_R, Spe is specificity, F1-mea1 is F1-measure, Mac\_F1 is Macro\_F1 and Var is variance. The red font value means that the value is the maximum value in the column (Unit: \%).}
\resizebox{\textwidth}{30mm}{
\begin{tabular}{|c|c|c|c|c|c|c|c|c|c|c|c|c|c|c|c|c|c|c|}
\hline
\textbf{Feature} & \textbf{Classifier}                                                              & \textbf{Acc}                                                               & \textbf{Pre1}                                                              & \textbf{Pre2}                                                              & \textbf{Pre3}                                                               & \textbf{Mac\_P}                                                            & \textbf{Rec1}                                                                & \textbf{Rec2}                                                                & \textbf{Rec3}                                                               & \textbf{Mac\_R}                                                            & \textbf{Spe1}                                                              & \textbf{Spe2}                                                              & \textbf{Spe3}                                                              & \textbf{F1-mea1}                                                           & \textbf{F1-mea2}                                                           & \textbf{F1-mea3}                                                           & \textbf{Mac\_F1}                                                           & \textbf{Var}                                                           \\ \hline
\textbf{HOG}     & \textbf{\begin{tabular}[c]{@{}c@{}}ANNs\\ RFs\\ RBF-SVM\\ Linear-SVM\end{tabular}} & \textbf{\begin{tabular}[c]{@{}c@{}}68.6\\ 71.2\\ 71.0\\ 71.0\end{tabular}} & \textbf{\begin{tabular}[c]{@{}c@{}}71.9\\ 72.6\\ 71.0\\ 71.0\end{tabular}} & \textbf{\begin{tabular}[c]{@{}c@{}}20.2\\ 43.5\\ 0.0\\ 0.0\end{tabular}}   & \textbf{\begin{tabular}[c]{@{}c@{}}35.1\\ 0.0\\ 0.0\\ 0.0\end{tabular}}     & \textbf{\begin{tabular}[c]{@{}c@{}}42.4\\ 38.7\\ 23.7\\ 23.7\end{tabular}} & \textbf{\begin{tabular}[c]{@{}c@{}}94.4\\ 97.5\\ {\color{red} 100.0}\\ {\color{red} 100.0}\end{tabular}} & \textbf{\begin{tabular}[c]{@{}c@{}}6.3\\ 10.7\\ 0.0\\ 0.0\end{tabular}}      & \textbf{\begin{tabular}[c]{@{}c@{}}3.5\\ 0.0\\ 0.0\\ 0.0\end{tabular}}      & \textbf{\begin{tabular}[c]{@{}c@{}}34.7\\ 36.1\\ 33.3\\ 33.3\end{tabular}} & \textbf{\begin{tabular}[c]{@{}c@{}}5.3\\ 6.7\\ 0.0\\ 0.0\end{tabular}}     & \textbf{\begin{tabular}[c]{@{}c@{}}82.3\\ 85.0\\ 86.7\\ 86.7\end{tabular}} & \textbf{\begin{tabular}[c]{@{}c@{}}76.5\\ 79.9\\ 79.7\\ 79.7\end{tabular}} & \textbf{\begin{tabular}[c]{@{}c@{}}81.6\\ 83.2\\ 83.0\\ 83.0\end{tabular}} & \textbf{\begin{tabular}[c]{@{}c@{}}9.6\\ 17.2\\ 0.0\\ 0.0\end{tabular}}    & \textbf{\begin{tabular}[c]{@{}c@{}}6.4\\ 0.0\\ 0.0\\ 0.0\end{tabular}}     & \textbf{\begin{tabular}[c]{@{}c@{}}38.2\\ 37.4\\ 27.8\\ 27.8\end{tabular}} & \textbf{\begin{tabular}[c]{@{}c@{}}0.51\\ 0.53\\ 0.58\\ 0.58\end{tabular}} \\ \hline
\textbf{GLCM}    & \textbf{\begin{tabular}[c]{@{}c@{}}ANNs\\ RFs\\ RBF-SVM\\ Linear-SVM\end{tabular}} & \textbf{\begin{tabular}[c]{@{}c@{}}69.7\\ 68.9\\ 18.0\\ 18.0\end{tabular}} & \textbf{\begin{tabular}[c]{@{}c@{}}75.9\\ 78.5\\ 0.0\\ 0.0\end{tabular}}   & \textbf{\begin{tabular}[c]{@{}c@{}}36.9\\ 31.4\\ 18.0\\ 18.0\end{tabular}} & \textbf{\begin{tabular}[c]{@{}c@{}}32.2\\ 31.6\\ 0.0\\ 0.0\end{tabular}}    & \textbf{\begin{tabular}[c]{@{}c@{}}48.3\\ 47.2\\ 6.0\\ 6.0\end{tabular}}   & \textbf{\begin{tabular}[c]{@{}c@{}}90.0\\ 88.0\\ 0.0\\ 0.0\end{tabular}}     & \textbf{\begin{tabular}[c]{@{}c@{}}30.6\\ 24.0\\ {\color{red} 100.0}\\ {\color{red} 100.0}\end{tabular}} & \textbf{\begin{tabular}[c]{@{}c@{}}25.3\\ 19.1\\ 0.0\\ 0.0\end{tabular}}    & \textbf{\begin{tabular}[c]{@{}c@{}}48.6\\ 43.7\\ 33.3\\ 33.3\end{tabular}} & \textbf{\begin{tabular}[c]{@{}c@{}}20.0\\ 22.3\\ 62.3\\ 62.3\end{tabular}} & \textbf{\begin{tabular}[c]{@{}c@{}}78.3\\ 78.8\\ 0.0\\ 0.0\end{tabular}}   & \textbf{\begin{tabular}[c]{@{}c@{}}77.9\\ 75.0\\ 20.3\\ 20.3\end{tabular}} & \textbf{\begin{tabular}[c]{@{}c@{}}82.4\\ 83.0\\ 0.0\\ 0.0\end{tabular}}   & \textbf{\begin{tabular}[c]{@{}c@{}}33.5\\ 27.2\\ 30.5\\ 30.5\end{tabular}} & \textbf{\begin{tabular}[c]{@{}c@{}}28.3\\ 23.8\\ 0.0\\ 0.0\end{tabular}}   & \textbf{\begin{tabular}[c]{@{}c@{}}48.4\\ 45.4\\ 10.2\\ 10.2\end{tabular}} & \textbf{\begin{tabular}[c]{@{}c@{}}0.36\\ 0.38\\ 0.58\\ 0.58\end{tabular}} \\ \hline
\textbf{HU}      & \textbf{\begin{tabular}[c]{@{}c@{}}ANNs\\ RFs\\ RBF-SVM\\ Linear-SVM\end{tabular}} & \textbf{\begin{tabular}[c]{@{}c@{}}71.0\\ 68.6\\ 71.0\\ 10.6\end{tabular}} & \textbf{\begin{tabular}[c]{@{}c@{}}71.0\\ 70.7\\ 71.0\\ 0.0\end{tabular}}  & \textbf{\begin{tabular}[c]{@{}c@{}}50.0\\ 15.3\\ 0.0\\ 0.0\end{tabular}}   & \textbf{\begin{tabular}[c]{@{}c@{}}0.0\\ 20.0\\ 0.0\\ 10.7\end{tabular}}    & \textbf{\begin{tabular}[c]{@{}c@{}}40.3\\ 35.3\\ 23.7\\ 3.6\end{tabular}}  & \textbf{\begin{tabular}[c]{@{}c@{}}99.98\\ 95.8\\ {\color{red} 100.0}\\ 0.0\end{tabular}}  & \textbf{\begin{tabular}[c]{@{}c@{}}0.08\\ 25.8\\ 0.0\\ 0.0\end{tabular}}     & \textbf{\begin{tabular}[c]{@{}c@{}}0.0\\ 13.3\\ 0.0\\ 97.3\end{tabular}}    & \textbf{\begin{tabular}[c]{@{}c@{}}33.4\\ 45.0\\ 33.3\\ 32.4\end{tabular}} & \textbf{\begin{tabular}[c]{@{}c@{}}0.05\\ 2.1\\ 0.0\\ 36.7\end{tabular}}   & \textbf{\begin{tabular}[c]{@{}c@{}}86.7\\ 83.2\\ 86.7\\ 13.0\end{tabular}} & \textbf{\begin{tabular}[c]{@{}c@{}}79.7\\ 76.9\\ 79.7\\ 0.0\end{tabular}}  & \textbf{\begin{tabular}[c]{@{}c@{}}83.0\\ 81.4\\ 83.0\\ 0.0\end{tabular}}  & \textbf{\begin{tabular}[c]{@{}c@{}}0.6\\ 4.4\\ 0.0\\ 0.0\end{tabular}}     & \textbf{\begin{tabular}[c]{@{}c@{}}0.0\\ 16.0\\ 0.0\\ 19.3\end{tabular}}   & \textbf{\begin{tabular}[c]{@{}c@{}}36.5\\ 39.6\\ 27.7\\ 6.5\end{tabular}}  & \textbf{\begin{tabular}[c]{@{}c@{}}0.58\\ 0.45\\ 0.58\\ 0.56\end{tabular}} \\ \hline
\textbf{SIFT}    & \textbf{\begin{tabular}[c]{@{}c@{}}ANNs\\ RFs\\ RBF-SVM\\ Linear-SVM\end{tabular}} & \textbf{\begin{tabular}[c]{@{}c@{}}70.8\\ 69.3\\ 71.0\\ 71.0\end{tabular}} & \textbf{\begin{tabular}[c]{@{}c@{}}71.0\\ 71.4\\ 71.0\\ 71.0\end{tabular}} & \textbf{\begin{tabular}[c]{@{}c@{}}20.0\\ 13.8\\ 0.0\\ 0.0\end{tabular}}   & \textbf{\begin{tabular}[c]{@{}c@{}}0.0\\ 33.3\\ 0.0\\ 0.0\end{tabular}}     & \textbf{\begin{tabular}[c]{@{}c@{}}30.3\\ 39.5\\ 23.7\\ 23.7\end{tabular}} & \textbf{\begin{tabular}[c]{@{}c@{}}99.7\\ 96.9\\ {\color{red} 100.0}\\ {\color{red} 100.0}\end{tabular}} & \textbf{\begin{tabular}[c]{@{}c@{}}0.3\\ 2.7\\ 0.0\\ 0.0\end{tabular}}       & \textbf{\begin{tabular}[c]{@{}c@{}}0.0\\ 0.1\\ 0.0\\ 0.0\end{tabular}}      & \textbf{\begin{tabular}[c]{@{}c@{}}33.3\\ 33.2\\ 33.3\\ 33.3\end{tabular}} & \textbf{\begin{tabular}[c]{@{}c@{}}0.2\\ 1.8\\ 0.0\\ 0.0\end{tabular}}     & \textbf{\begin{tabular}[c]{@{}c@{}}86.4\\ 84.0\\ 86.7\\ 86.7\end{tabular}} & \textbf{\begin{tabular}[c]{@{}c@{}}79.5\\ 77.8\\ 79.7\\ 79.7\end{tabular}} & \textbf{\begin{tabular}[c]{@{}c@{}}82.9\\ 82.2\\ 83.0\\ 83.0\end{tabular}} & \textbf{\begin{tabular}[c]{@{}c@{}}0.6\\ 4.5\\ 0.0\\ 0.0\end{tabular}}     & \textbf{\begin{tabular}[c]{@{}c@{}}0.0\\ 0.2\\ 0.0\\ 0.0\end{tabular}}     & \textbf{\begin{tabular}[c]{@{}c@{}}31.7\\ 36.0\\ 27.7\\ 27.7\end{tabular}} & \textbf{\begin{tabular}[c]{@{}c@{}}0.58\\ 0.55\\ 0.58\\ 0.58\end{tabular}} \\ \hline
\textbf{GRAY}    & \textbf{\begin{tabular}[c]{@{}c@{}}ANNs\\ RFs\\ RBF-SVM\\ Linear-SVM\end{tabular}} & \textbf{\begin{tabular}[c]{@{}c@{}}69.0\\ 71.0\\ 71.0\\ 71.0\end{tabular}} & \textbf{\begin{tabular}[c]{@{}c@{}}71.6\\ 71.3\\ 71.0\\ 71.0\end{tabular}} & \textbf{\begin{tabular}[c]{@{}c@{}}21.8\\ 35.4\\ 0.0\\ 0.0\end{tabular}}   & \textbf{\begin{tabular}[c]{@{}c@{}}9.5\\ 40.0\\ 0.0\\ 0.0\end{tabular}}           & \textbf{\begin{tabular}[c]{@{}c@{}}34.3\\ 48.9\\ 23.7\\ 23.7\end{tabular}} & \textbf{\begin{tabular}[c]{@{}c@{}}95.7\\ 99.4\\ 99.4\\ {\color{red} 100.0}\end{tabular}}  & \textbf{\begin{tabular}[c]{@{}c@{}}5.7\\ 1.4\\ 0.0\\ 0.0\end{tabular}}       & \textbf{\begin{tabular}[c]{@{}c@{}}0.3\\ 1.1\\ 0.0\\ 0.0\end{tabular}}      & \textbf{\begin{tabular}[c]{@{}c@{}}33.9\\ 34.0\\ 33.1\\ 33.3\end{tabular}} & \textbf{\begin{tabular}[c]{@{}c@{}}3.8\\ 1.3\\ 0.0\\ 0.0\end{tabular}}     & \textbf{\begin{tabular}[c]{@{}c@{}}83.0\\ 86.3\\ 86.1\\ 86.7\end{tabular}} & \textbf{\begin{tabular}[c]{@{}c@{}}77.4\\ 79.5\\ 79.3\\ 79.7\end{tabular}} & \textbf{\begin{tabular}[c]{@{}c@{}}81.9\\ 83.0\\ 82.9\\ 83.0\end{tabular}} & \textbf{\begin{tabular}[c]{@{}c@{}}9.0\\ 2.7\\ 0.0\\ 0.0\end{tabular}}     & \textbf{\begin{tabular}[c]{@{}c@{}}0.6\\ 2.1\\ 0.0\\ 0.0\end{tabular}}     & \textbf{\begin{tabular}[c]{@{}c@{}}34.1\\ 40.1\\ 27.6\\ 27.7\end{tabular}} & \textbf{\begin{tabular}[c]{@{}c@{}}0.54\\ 0.57\\ 0.57\\ 0.57\end{tabular}} \\ \hline
\textbf{$F_{\left( i,g \right)}^{\text{X}}$}       & \textbf{\begin{tabular}[c]{@{}c@{}}ANNs\\ RFs\\ RBF-SVM\\ Linear-SVM\end{tabular}} & \textbf{\begin{tabular}[c]{@{}c@{}}77.1\\ 76.4\\ 73.2\\ 76.7\end{tabular}} & \textbf{\begin{tabular}[c]{@{}c@{}}81.2\\ 84.3\\ 74.2\\ 79.8\end{tabular}} & \textbf{\begin{tabular}[c]{@{}c@{}}50.7\\ 46.3\\ 52.9\\ 51.9\end{tabular}} & \textbf{\begin{tabular}[c]{@{}c@{}}64.7\\ 47.7\\ {\color{red} 100.0}\\ 61.1\end{tabular}} & \textbf{\begin{tabular}[c]{@{}c@{}}65.5\\ 59.4\\ 75.7\\ 64.3\end{tabular}} & \textbf{\begin{tabular}[c]{@{}c@{}}96.5\\ 93.6\\ 99.0\\ 98.0\end{tabular}}   & \textbf{\begin{tabular}[c]{@{}c@{}}29.8\\ 29.8\\ 14.5\\ 21.8\end{tabular}}   & \textbf{\begin{tabular}[c]{@{}c@{}}29.3\\ 41.3\\ 2.7\\ 29.3\end{tabular}}   & \textbf{\begin{tabular}[c]{@{}c@{}}51.9\\ 54.9\\ 38.7\\ 49.7\end{tabular}} & \textbf{\begin{tabular}[c]{@{}c@{}}29.7\\ 34.2\\ 10.0\\ 24.6\end{tabular}} & \textbf{\begin{tabular}[c]{@{}c@{}}87.6\\ 86.7\\ 86.4\\ 88.8\end{tabular}} & \textbf{\begin{tabular}[c]{@{}c@{}}83.0\\ 80.7\\ 81.9\\ 82.5\end{tabular}} & \textbf{\begin{tabular}[c]{@{}c@{}}88.2\\ 88.7\\ 84.8\\ 88.0\end{tabular}} & \textbf{\begin{tabular}[c]{@{}c@{}}37.5\\ 36.3\\ 22.8\\ 30.7\end{tabular}} & \textbf{\begin{tabular}[c]{@{}c@{}}40.3\\ 44.3\\ 5.3\\ 39.6\end{tabular}}  & \textbf{\begin{tabular}[c]{@{}c@{}}57.9\\ 7.0\\ 51.2\\ 56.0\end{tabular}}  & \textbf{\begin{tabular}[c]{@{}c@{}}0.39\\ 0.34\\ 0.52\\ 0.42\end{tabular}} \\ \hline
\textbf{$F_{\left( i,g \right)}^{\text{Y}}$}       & \textbf{\begin{tabular}[c]{@{}c@{}}ANNs\\ RFs\\ RBF-SVM\\ Linear-SVM\end{tabular}} & \textbf{\begin{tabular}[c]{@{}c@{}}82.7\\ 83.0\\ 73.8\\ 77.7\end{tabular}} & \textbf{\begin{tabular}[c]{@{}c@{}}94.9\\ 91.7\\ 74.4\\ 95.2\end{tabular}} & \textbf{\begin{tabular}[c]{@{}c@{}}53.6\\ 58.0\\ 0.0\\ 28.6\end{tabular}}  & \textbf{\begin{tabular}[c]{@{}c@{}}64.3\\ 63.9\\ 62.5\\ 37.5\end{tabular}}  & \textbf{\begin{tabular}[c]{@{}c@{}}70.9\\ 71.2\\ 45.6\\ 53.8\end{tabular}} & \textbf{\begin{tabular}[c]{@{}c@{}}90.1\\ 93.2\\ 99.8\\ 93.6\end{tabular}}   & \textbf{\begin{tabular}[c]{@{}c@{}}65.3\\ 55.6\\ 0.0\\ 1.6\end{tabular}}     & \textbf{\begin{tabular}[c]{@{}c@{}}60.0\\ 61.3\\ 26.7\\ {\color{red} 100.0}\end{tabular}} & \textbf{\begin{tabular}[c]{@{}c@{}}71.8\\ 70.0\\ 42.2\\ 65.1\end{tabular}} & \textbf{\begin{tabular}[c]{@{}c@{}}63.3\\ 57.8\\ 10.0\\ 38.7\end{tabular}} & \textbf{\begin{tabular}[c]{@{}c@{}}86.5\\ 89.0\\ 90.0\\ 94.5\end{tabular}} & \textbf{\begin{tabular}[c]{@{}c@{}}85.5\\ 85.6\\ 79.6\\ 75.0\end{tabular}} & \textbf{\begin{tabular}[c]{@{}c@{}}92.4\\ 92.4\\ 85.2\\ 94.4\end{tabular}} & \textbf{\begin{tabular}[c]{@{}c@{}}58.9\\ 56.8\\ 0.0\\ 3.0\end{tabular}}   & \textbf{\begin{tabular}[c]{@{}c@{}}62.1\\ 62.6\\ 37.4\\ 54.5\end{tabular}} & \textbf{\begin{tabular}[c]{@{}c@{}}71.3\\ 70.1\\ 43.8\\ 58.9\end{tabular}} & \textbf{\begin{tabular}[c]{@{}c@{}}0.16\\ 0.20\\ 0.52\\ 0.55\end{tabular}} \\ \hline
\textbf{$F_{\left( i,g \right)}^{\text{Z}}$}       & \textbf{\begin{tabular}[c]{@{}c@{}}ANNs\\ RFs\\ RBF-SVM\\ Linear-SVM\end{tabular}} & \textbf{\begin{tabular}[c]{@{}c@{}}{\color{red} 91.8}\\ 85.7\\ 79.5\\ 80.1\end{tabular}} & \textbf{\begin{tabular}[c]{@{}c@{}}{\color{red} 98.7}\\ 92.8\\ 90.9\\ 98.3\end{tabular}} & \textbf{\begin{tabular}[c]{@{}c@{}}{\color{red} 73.0}\\ 62.7\\ 62.5\\ 51.5\end{tabular}} & \textbf{\begin{tabular}[c]{@{}c@{}}{\color{red} 86.8}\\ 83.8\\ 43.5\\ 63.6\end{tabular}}  & \textbf{\begin{tabular}[c]{@{}c@{}}{\color{red} 86.2}\\ 80.0\\ 65.6\\ 71.1\end{tabular}} & \textbf{\begin{tabular}[c]{@{}c@{}}93.5\\ 95.0\\ 96.5\\ 82.6\end{tabular}}   & \textbf{\begin{tabular}[c]{@{}c@{}}87.1\\ 75.8\\ 4.0\\ 83.1\end{tabular}}    & \textbf{\begin{tabular}[c]{@{}c@{}}89.2\\ 41.3\\ 93.3\\ 65.3\end{tabular}}  & \textbf{\begin{tabular}[c]{@{}c@{}}{\color{red} 89.9}\\ 70.7\\ 64.6\\ 77.0\end{tabular}} & \textbf{\begin{tabular}[c]{@{}c@{}}{\color{red} 87.9}\\ 62.8\\ 37.7\\ 76.4\end{tabular}} & \textbf{\begin{tabular}[c]{@{}c@{}}92.9\\ 87.9\\ {\color{red} 96.1}\\ 80.3\end{tabular}} & \textbf{\begin{tabular}[c]{@{}c@{}}{\color{red} 92.3}\\ 91.2\\ 77.8\\ 82.7\end{tabular}} & \textbf{\begin{tabular}[c]{@{}c@{}}{\color{red} 96.0}\\ 93.4\\ 93.6\\ 89.8\end{tabular}} & \textbf{\begin{tabular}[c]{@{}c@{}}{\color{red} 79.4}\\ 68.6\\ 7.5\\ 63.6\end{tabular}}  & \textbf{\begin{tabular}[c]{@{}c@{}}{\color{red} 88.0}\\ 55.3\\ 59.3\\ 64.4\end{tabular}} & \textbf{\begin{tabular}[c]{@{}c@{}}{\color{red} 88.0}\\ 75.0\\ 65.1\\ 73.9\end{tabular}} & \textbf{\begin{tabular}[c]{@{}c@{}}{\color{red} 0.04}\\ 0.27\\ 0.53\\ 0.10 \end{tabular}} \\ \hline
\end{tabular}}
\label{tab: static features and foldover features}
\end{table*}

Considering the comparison in TABLE\ref{tab: static features and foldover features}, the accuracy of $F_{\left( i,g \right)}^{\text{X}}$, $F_{\left( i,g \right)}^{\text{Y}}$ and $F_{\left( i,g \right)}^{\text{Z}}$ are significantly higher than that of static features. The reason for the low accuracy of static features is: Static features are extracted from detected sperms images, in which there is few difference between stationary sperms and moving sperms, therefore it is difficult to distinguish different categories of sperms by static features. Because there are not many differences between static sperms and moving sperms, it is easy to miss-classify all sperms into one category by using static features to classify sperm in different motion states, consequently, one precision (precision 1, precision 2 and precision 3) for one static feature is very high and the others are very low. The case for recall values are totally similar to that of the precision. The difference of values between precision 1, precision 2 and precision 3 further affect the macro of precision, recall and F1-measure. 

$F_{\left( i,g \right)}^{\text{X}}$, $F_{\left( i,g \right)}^{\text{Y}}$ and $F_{\left( i,g \right)}^{\text{Z}}$ can distinguish three categories well by the advantage of foldover in classification, especially the information of foldover in the Z direction is very beneficial to distinguish sperms in different motion states. Therefore, the value of precision and recall are higher than which of the static features. Furthermore, foldover features perform well in the macro of precision, recall and F1-measure.

\subsubsection{Comparison with Dynamic Features}
\begin{figure*}[htbp]
\centering
\includegraphics[scale=0.19]{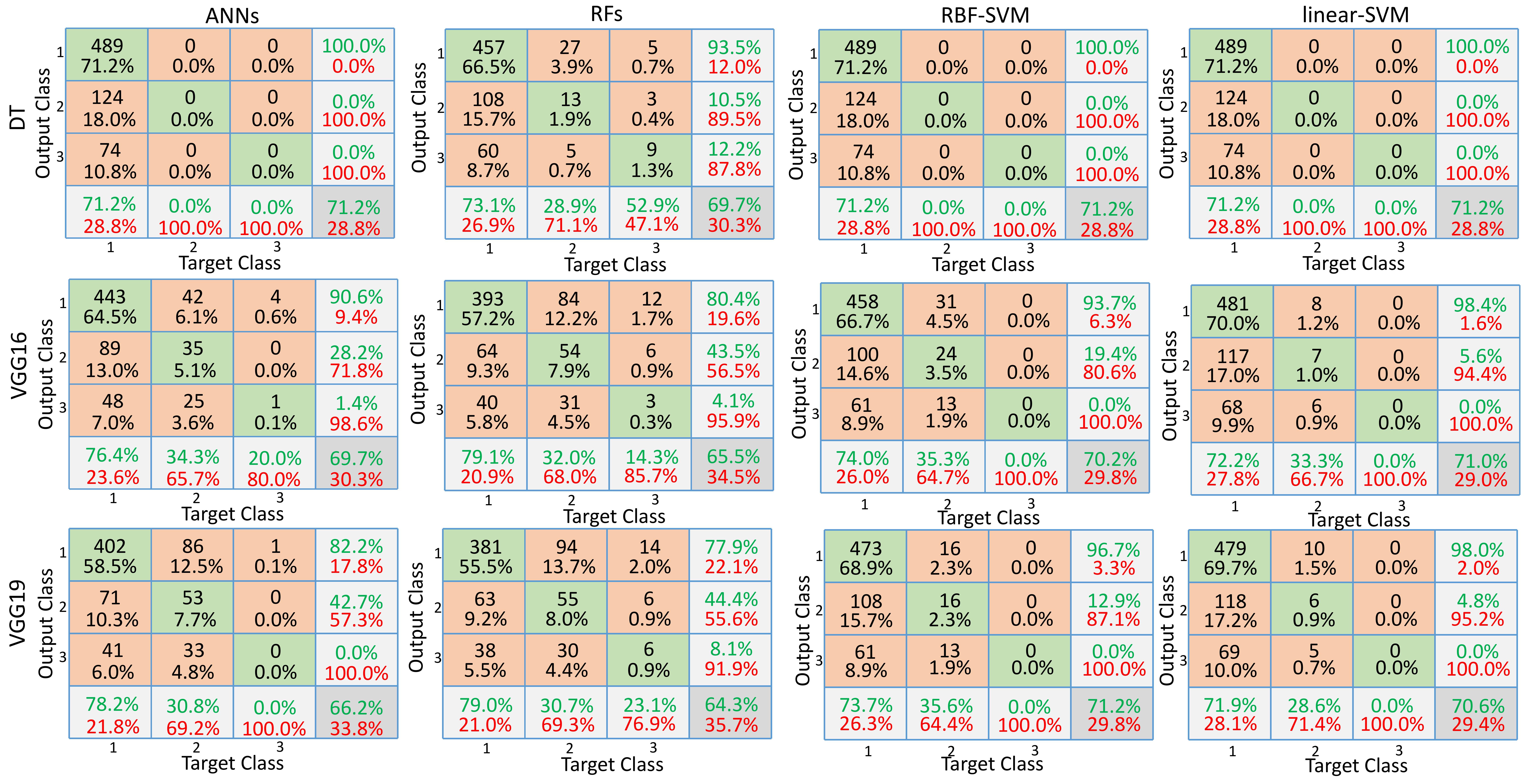}
\caption{The confusion matrices of dynamic features. Rows represent feature types, columns represent classifier types, and DT represents dynamic texture features.}
\label{fig:dynamic_feature_confusion_matrix}
\end{figure*}

Three dynamic features are selected for the comparative experiment, including dynamic texture features and features extracted based on the CNNs (VGG-16 and VGG-19 networks). The first step is the same as the operation of static features, where each sperm is detected to a size of 26 by 26 pixels in the corresponding frame. The difference is what we need is the entire movement of the detected sperm. Therefore, the detected sperm images are combined into a video of the corresponding sperm. Secondly, we refer to the articles~\cite{Soatto-2001-DT}~\cite{Wang-2016-Visual}~\cite{Chao-2015-Hierarchical} to extract dynamic texture and deep learning (VGG-16 and VGG-19 networks) features in the detected semen videos. The third step is the same as the operation of static features, where we use ANNs~\cite{Judith-ANN-2001}, RFs~\cite{Ho-RF-1995} and SVM~\cite{Saunders-SVM-2002} (Linear- and RBF-SVM) classifiers to distinguish dynamic features, and the classification results are shown in \figurename~\ref{fig:dynamic_feature_confusion_matrix}.

According to the classification results of dynamic features in \figurename~\ref{fig:dynamic_feature_confusion_matrix}, we compare evaluations between dynamic features and foldover features in TABLE\ref{tab: dynamic features and foldover features}.

\begin{table*}[htbp]
\tiny
\caption{Evaluation of dynamic features with four classifiers. The first column shows the types of dynamic features, the second column shows the types of classifiers, the third to the last columns show the calculated evaluations. We use the first three letters of each evaluation to indicate the evaluation metric, such as Acc is accuracy, Pre is precision, Mac\_P is Macro\_P, Rec is recall, Mac\_R is Macro\_R, Spe is specificity, F1-mea1 is F1-measure, Mac\_F1 is Macro\_F1 and Var is variance. The red font value means that the value is the maximum value in the column (Unit: \%).}
\resizebox{\textwidth}{30mm}{
\begin{tabular}{|c|c|c|c|c|c|c|c|c|c|c|c|c|c|c|c|c|c|c|}
\hline
\textbf{Feature} & \textbf{Classifier}                                                              & \textbf{Acc}                                                               & \textbf{Pre1}                                                              & \textbf{Pre2}                                                              & \textbf{Pre3}                                                               & \textbf{Mac\_P}                                                            & \textbf{Rec1}                                                                 & \textbf{Rec2}                                                              & \textbf{Rec3}                                                               & \textbf{Mac\_R}                                                            & \textbf{Spe1}                                                              & \textbf{Spe2}                                                              & \textbf{Spe3}                                                              & \textbf{F1-mea1}                                                           & \textbf{F1-mea2}                                                           & \textbf{F1-mea3}                                                           & \textbf{Mac\_F1}                                                           & \textbf{Var}                                                           \\ \hline
\textbf{DT}      & \textbf{\begin{tabular}[c]{@{}c@{}}ANNs\\ RFs\\ RBF-SVM\\ Linear-SVM\end{tabular}} & \textbf{\begin{tabular}[c]{@{}c@{}}71.2\\ 69.7\\ 71.2\\ 71.2\end{tabular}} & \textbf{\begin{tabular}[c]{@{}c@{}}71.2\\ 73.1\\ 71.2\\ 71.2\end{tabular}} & \textbf{\begin{tabular}[c]{@{}c@{}}0.0\\ 28.9\\ 0.0\\ 0.0\end{tabular}}    & \textbf{\begin{tabular}[c]{@{}c@{}}0.0\\ 52.9\\ 0.0\\ 0.0\end{tabular}}     & \textbf{\begin{tabular}[c]{@{}c@{}}23.7\\ 51.6\\ 23.7\\ 23.7\end{tabular}} & \textbf{\begin{tabular}[c]{@{}c@{}}{\color{red} 100.0}\\ 93.5\\ {\color{red} 100.0}\\ {\color{red} 100.0}\end{tabular}} & \textbf{\begin{tabular}[c]{@{}c@{}}0.0\\ 10.5\\ 0.0\\ 0.0\end{tabular}}    & \textbf{\begin{tabular}[c]{@{}c@{}}0.0\\ 12.2\\ 0.0\\ 0.0\end{tabular}}     & \textbf{\begin{tabular}[c]{@{}c@{}}33.3\\ 38.7\\ 33.3\\ 33.3\end{tabular}} & \textbf{\begin{tabular}[c]{@{}c@{}}0.0\\ 10.0\\ 0.0\\ 0.0\end{tabular}}    & \textbf{\begin{tabular}[c]{@{}c@{}}86.9\\ 82.3\\ 86.9\\ 86.9\end{tabular}} & \textbf{\begin{tabular}[c]{@{}c@{}}79.8\\ 76.7\\ 79.8\\ 79.8\end{tabular}} & \textbf{\begin{tabular}[c]{@{}c@{}}83.2\\ 82.1\\ 83.2\\ 83.2\end{tabular}} & \textbf{\begin{tabular}[c]{@{}c@{}}0.0\\ 15.4\\ 0.0\\ 0.0\end{tabular}}    & \textbf{\begin{tabular}[c]{@{}c@{}}0.0\\ 19.8\\ 0.0\\ 0.0\end{tabular}}    & \textbf{\begin{tabular}[c]{@{}c@{}}27.7\\ 44.2\\ 27.7\\ 27.7\end{tabular}} & \textbf{\begin{tabular}[c]{@{}c@{}}0.58\\ 0.48\\ 0.58\\ 0.58\end{tabular}} \\ \hline
\textbf{VGG16}   & \textbf{\begin{tabular}[c]{@{}c@{}}ANNs\\ RFs\\ RBF-SVM\\ Linear-SVM\end{tabular}} & \textbf{\begin{tabular}[c]{@{}c@{}}69.7\\ 65.5\\ 70.2\\ 71.0\end{tabular}} & \textbf{\begin{tabular}[c]{@{}c@{}}76.4\\ 79.1\\ 74.0\\ 72.2\end{tabular}} & \textbf{\begin{tabular}[c]{@{}c@{}}34.3\\ 32.0\\ 35.3\\ 33.3\end{tabular}} & \textbf{\begin{tabular}[c]{@{}c@{}}20.0\\ 14.3\\ 0.0\\ 0.0\end{tabular}}    & \textbf{\begin{tabular}[c]{@{}c@{}}43.6\\ 41.8\\ 36.4\\ 35.2\end{tabular}} & \textbf{\begin{tabular}[c]{@{}c@{}}90.6\\ 80.4\\ 93.7\\ 98.4\end{tabular}}    & \textbf{\begin{tabular}[c]{@{}c@{}}28.2\\ 43.5\\ 19.4\\ 5.6\end{tabular}}  & \textbf{\begin{tabular}[c]{@{}c@{}}1.4\\ 4.1\\ 0.0\\ 0.0\end{tabular}}      & \textbf{\begin{tabular}[c]{@{}c@{}}40.1\\ 42.7\\ 37.7\\ 34.7\end{tabular}} & \textbf{\begin{tabular}[c]{@{}c@{}}18.2\\ 28.8\\ 12.1\\ 35.4\end{tabular}} & \textbf{\begin{tabular}[c]{@{}c@{}}78.9\\ 70.3\\ 81.4\\ 76.1\end{tabular}} & \textbf{\begin{tabular}[c]{@{}c@{}}78.0\\ 72.9\\ 78.6\\ 79.6\end{tabular}} & \textbf{\begin{tabular}[c]{@{}c@{}}82.9\\ 79.7\\ 82.7\\ 83.3\end{tabular}} & \textbf{\begin{tabular}[c]{@{}c@{}}31.0\\ 36.9\\ 25.0\\ 9.6\end{tabular}}  & \textbf{\begin{tabular}[c]{@{}c@{}}2.6\\ 6.4\\ 0.0\\ 0.0\end{tabular}}     & \textbf{\begin{tabular}[c]{@{}c@{}}41.8\\ 42.2\\ 37.0\\ 34.9\end{tabular}} & \textbf{\begin{tabular}[c]{@{}c@{}}0.46\\ 0.38\\ 0.50\\ 0.55\end{tabular}} \\ \hline
\textbf{VGG19}   & \textbf{\begin{tabular}[c]{@{}c@{}}ANNs\\ RFs\\ RBF-SVM\\ Linear-SVM\end{tabular}} & \textbf{\begin{tabular}[c]{@{}c@{}}66.2\\ 64.3\\ 71.2\\ 70.6\end{tabular}} & \textbf{\begin{tabular}[c]{@{}c@{}}78.2\\ 79.0\\ 73.7\\ 71.9\end{tabular}} & \textbf{\begin{tabular}[c]{@{}c@{}}30.8\\ 30.7\\ 35.6\\ 28.6\end{tabular}} & \textbf{\begin{tabular}[c]{@{}c@{}}0.0\\ 23.1\\ 0.0\\ 0.0\end{tabular}}     & \textbf{\begin{tabular}[c]{@{}c@{}}36.3\\ 44.3\\ 36.4\\ 33.5\end{tabular}} & \textbf{\begin{tabular}[c]{@{}c@{}}82.2\\ 77.9\\ 96.7\\ 98.0\end{tabular}}    & \textbf{\begin{tabular}[c]{@{}c@{}}42.7\\ 44.4\\ 12.9\\ 4.8\end{tabular}}  & \textbf{\begin{tabular}[c]{@{}c@{}}0.0\\ 8.1\\ 0.0\\ 0.0\end{tabular}}      & \textbf{\begin{tabular}[c]{@{}c@{}}41.5\\ 43.5\\ 36.5\\ 34.3\end{tabular}} & \textbf{\begin{tabular}[c]{@{}c@{}}26.8\\ 30.8\\ 8.1\\ 3.0\end{tabular}}   & \textbf{\begin{tabular}[c]{@{}c@{}}71.4\\ 68.7\\ 84.0\\ 85.1\end{tabular}} & \textbf{\begin{tabular}[c]{@{}c@{}}74.2\\ 71.1\\ 79.8\\ 79.1\end{tabular}} & \textbf{\begin{tabular}[c]{@{}c@{}}80.2\\ 78.4\\ 83.6\\ 82.9\end{tabular}} & \textbf{\begin{tabular}[c]{@{}c@{}}35.8\\ 36.3\\ 18.9\\ 8.2\end{tabular}}  & \textbf{\begin{tabular}[c]{@{}c@{}}0.0\\ 12.0\\ 0.0\\ 0.0\end{tabular}}    & \textbf{\begin{tabular}[c]{@{}c@{}}38.7\\ 43.9\\ 36.4\\ 33.9\end{tabular}} & \textbf{\begin{tabular}[c]{@{}c@{}}0.41\\ 0.35\\ 0.53\\ 0.55\end{tabular}} \\ \hline
\textbf{$F_{\left( i,g \right)}^{\text{X}}$}       & \textbf{\begin{tabular}[c]{@{}c@{}}ANNs\\ RFs\\ RBF-SVM\\ Linear-SVM\end{tabular}} & \textbf{\begin{tabular}[c]{@{}c@{}}77.1\\ 76.4\\ 73.2\\ 76.7\end{tabular}} & \textbf{\begin{tabular}[c]{@{}c@{}}81.2\\ 84.3\\ 74.2\\ 79.8\end{tabular}} & \textbf{\begin{tabular}[c]{@{}c@{}}50.7\\ 46.3\\ 52.9\\ 51.9\end{tabular}} & \textbf{\begin{tabular}[c]{@{}c@{}}64.7\\ 47.7\\ {\color{red} 100.0}\\ 61.1\end{tabular}} & \textbf{\begin{tabular}[c]{@{}c@{}}65.5\\ 59.4\\ 75.7\\ 64.3\end{tabular}} & \textbf{\begin{tabular}[c]{@{}c@{}}96.5\\ 93.6\\ 99.0\\ 98.0\end{tabular}}    & \textbf{\begin{tabular}[c]{@{}c@{}}29.8\\ 29.8\\ 14.5\\ 21.8\end{tabular}} & \textbf{\begin{tabular}[c]{@{}c@{}}29.3\\ 41.3\\ 2.7\\ 29.3\end{tabular}}   & \textbf{\begin{tabular}[c]{@{}c@{}}51.9\\ 54.9\\ 38.7\\ 49.7\end{tabular}} & \textbf{\begin{tabular}[c]{@{}c@{}}29.7\\ 34.2\\ 10.0\\ 24.6\end{tabular}} & \textbf{\begin{tabular}[c]{@{}c@{}}87.6\\ 86.7\\ 86.4\\ 88.8\end{tabular}} & \textbf{\begin{tabular}[c]{@{}c@{}}83.0\\ 80.7\\ 81.9\\ 82.5\end{tabular}} & \textbf{\begin{tabular}[c]{@{}c@{}}88.2\\ 88.7\\ 84.8\\ 88.0\end{tabular}} & \textbf{\begin{tabular}[c]{@{}c@{}}37.5\\ 36.3\\ 22.8\\ 30.7\end{tabular}} & \textbf{\begin{tabular}[c]{@{}c@{}}40.3\\ 44.3\\ 5.3\\ 39.6\end{tabular}}  & \textbf{\begin{tabular}[c]{@{}c@{}}57.9\\ 7.0\\ 51.2\\ 56.0\end{tabular}}  & \textbf{\begin{tabular}[c]{@{}c@{}}0.39\\ 0.34\\ 0.52\\ 0.42\end{tabular}} \\ \hline
\textbf{$F_{\left( i,g \right)}^{\text{Y}}$}       & \textbf{\begin{tabular}[c]{@{}c@{}}ANNs\\ RFs\\ RBF-SVM\\ Linear-SVM\end{tabular}} & \textbf{\begin{tabular}[c]{@{}c@{}}82.7\\ 83.0\\ 73.8\\ 77.7\end{tabular}} & \textbf{\begin{tabular}[c]{@{}c@{}}94.9\\ 91.7\\ 74.4\\ 95.2\end{tabular}} & \textbf{\begin{tabular}[c]{@{}c@{}}53.6\\ 58.0\\ 0.0\\ 28.6\end{tabular}}  & \textbf{\begin{tabular}[c]{@{}c@{}}64.3\\ 63.9\\ 62.5\\ 37.5\end{tabular}}  & \textbf{\begin{tabular}[c]{@{}c@{}}70.9\\ 71.2\\ 45.6\\ 53.8\end{tabular}} & \textbf{\begin{tabular}[c]{@{}c@{}}90.1\\ 93.2\\ 99.8\\ 93.6\end{tabular}}    & \textbf{\begin{tabular}[c]{@{}c@{}}65.3\\ 55.6\\ 0.0\\ 1.6\end{tabular}}   & \textbf{\begin{tabular}[c]{@{}c@{}}60.0\\ 61.3\\ 26.7\\ {\color{red} 100.0}\end{tabular}} & \textbf{\begin{tabular}[c]{@{}c@{}}71.8\\ 70.0\\ 42.2\\ 65.1\end{tabular}} & \textbf{\begin{tabular}[c]{@{}c@{}}63.3\\ 57.8\\ 10.0\\ 38.7\end{tabular}} & \textbf{\begin{tabular}[c]{@{}c@{}}86.5\\ 89.0\\ 90.0\\ 94.5\end{tabular}} & \textbf{\begin{tabular}[c]{@{}c@{}}85.5\\ 85.6\\ 79.6\\ 75.0\end{tabular}} & \textbf{\begin{tabular}[c]{@{}c@{}}92.4\\ 92.4\\ 85.2\\ 94.4\end{tabular}} & \textbf{\begin{tabular}[c]{@{}c@{}}58.9\\ 56.8\\ 0.0\\ 3.0\end{tabular}}   & \textbf{\begin{tabular}[c]{@{}c@{}}62.1\\ 62.6\\ 37.4\\ 54.5\end{tabular}} & \textbf{\begin{tabular}[c]{@{}c@{}}71.3\\ 70.1\\ 43.8\\ 58.9\end{tabular}} & \textbf{\begin{tabular}[c]{@{}c@{}}0.16\\ 0.20\\ 0.52\\ 0.55\end{tabular}} \\ \hline
\textbf{$F_{\left( i,g \right)}^{\text{Z}}$}       & \textbf{\begin{tabular}[c]{@{}c@{}}ANNs\\ RFs\\ RBF-SVM\\ Linear-SVM\end{tabular}} & \textbf{\begin{tabular}[c]{@{}c@{}}{\color{red} 91.8}\\ 85.7\\ 79.5\\ 80.1\end{tabular}} & \textbf{\begin{tabular}[c]{@{}c@{}}{\color{red} 98.7}\\ 92.8\\ 90.9\\ 98.3\end{tabular}} & \textbf{\begin{tabular}[c]{@{}c@{}}{\color{red} 73.0}\\ 62.7\\ 62.5\\ 51.5\end{tabular}} & \textbf{\begin{tabular}[c]{@{}c@{}}86.8\\ 83.8\\ 43.5\\ 63.6\end{tabular}}  & \textbf{\begin{tabular}[c]{@{}c@{}}{\color{red} 86.2}\\ 80.0\\ 65.6\\ 71.1\end{tabular}} & \textbf{\begin{tabular}[c]{@{}c@{}}93.5\\ 95.0\\ 96.5\\ 82.6\end{tabular}}    & \textbf{\begin{tabular}[c]{@{}c@{}}{\color{red} 87.1}\\ 75.8\\ 4.0\\ 83.1\end{tabular}}  & \textbf{\begin{tabular}[c]{@{}c@{}}89.2\\ 41.3\\ 93.3\\ 65.3\end{tabular}}  & \textbf{\begin{tabular}[c]{@{}c@{}}{\color{red} 89.9}\\ 70.7\\ 64.6\\ 77.0\end{tabular}} & \textbf{\begin{tabular}[c]{@{}c@{}}{\color{red} 87.9}\\ 62.8\\ 37.7\\ 76.4\end{tabular}} & \textbf{\begin{tabular}[c]{@{}c@{}}92.9\\ 87.9\\ {\color{red} 96.1}\\ 80.3\end{tabular}} & \textbf{\begin{tabular}[c]{@{}c@{}}{\color{red} 92.3}\\ 91.2\\ 77.8\\ 82.7\end{tabular}} & \textbf{\begin{tabular}[c]{@{}c@{}}{\color{red} 96.0}\\ 93.4\\ 93.6\\ 89.8\end{tabular}} & \textbf{\begin{tabular}[c]{@{}c@{}}{\color{red} 79.4}\\ 68.6\\ 7.5\\ 63.6\end{tabular}}  & \textbf{\begin{tabular}[c]{@{}c@{}}{\color{red} 88.0}\\ 55.3\\ 59.3\\ 64.4\end{tabular}} & \textbf{\begin{tabular}[c]{@{}c@{}}{\color{red} 88.0}\\ 75.0\\ 65.1\\ 73.9\end{tabular}} & \textbf{\begin{tabular}[c]{@{}c@{}}{\color{red} 0.04}\\ 0.27\\ 0.53\\ 0.10\end{tabular}} \\ \hline
\end{tabular}}
\label{tab: dynamic features and foldover features}
\end{table*}

Considering the comparison in TABLE\ref{tab: dynamic features and foldover features}, the accuracy of $F_{\left( i,g \right)}^{\text{X}}$, $F_{\left( i,g \right)}^{\text{Y}}$ and $F_{\left( i,g \right)}^{\text{Z}}$ are significantly higher than that of dynamic features. The reason for the low accuracy of dynamic features is: Sperms are tiny and there is very little dynamic information. Therefore, it is difficult to distinguish different categories of sperms by dynamic features. It is easy to classify most of sperms into one category by using dynamic features to classify sperm in different motion states, consequently, the value of the true positive (TP) further affect the calculation results of all evaluations.

\subsubsection{Experimental Analysis}
\begin{figure}[H]
\centering
\includegraphics[scale=0.7]{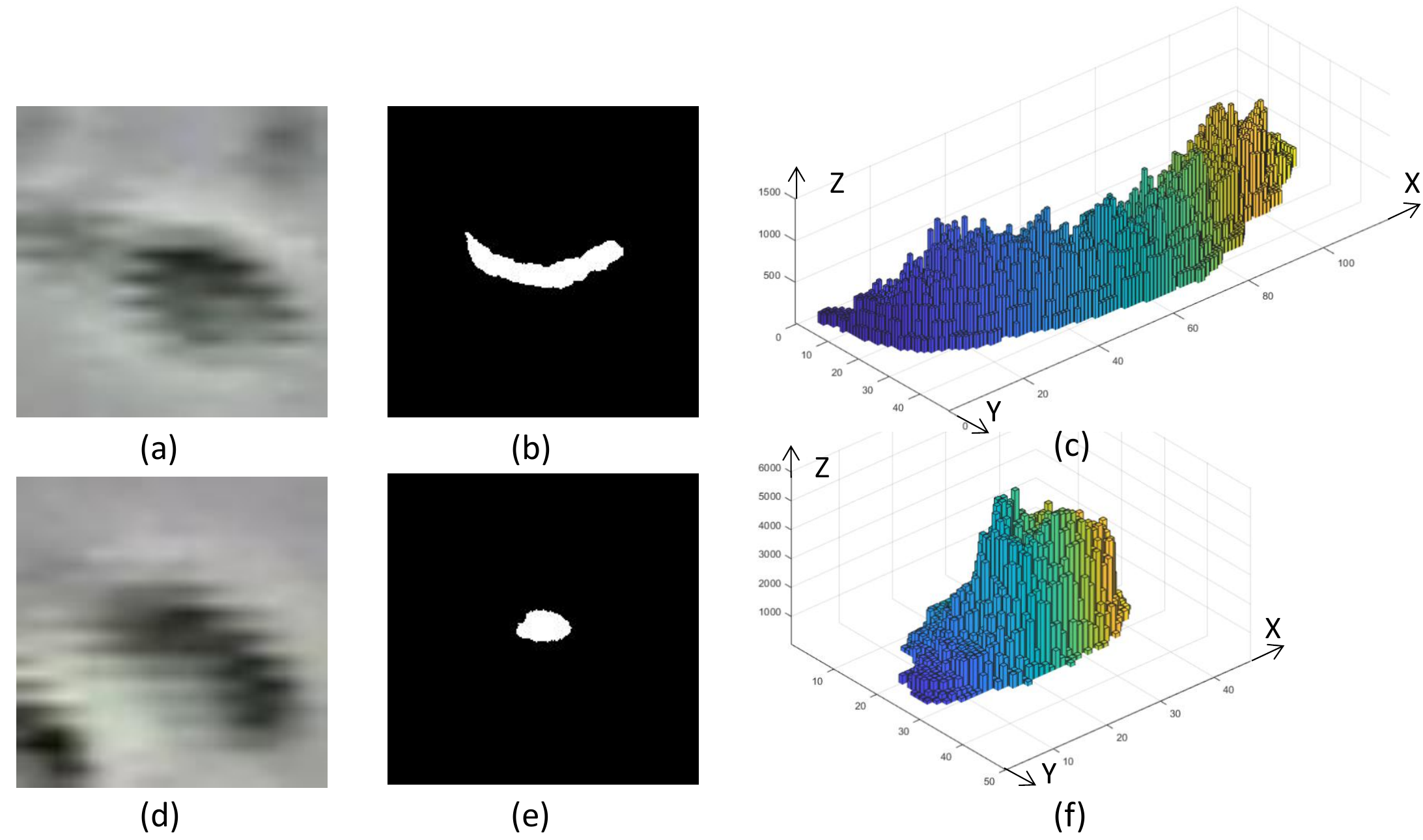}
\caption{An Example of different sperms. (a) and (d) represent two different sperms, (b) is the foldover of (a) in the Z direction,  (e) is the foldover of (d) in the Z direction, (c) is the foldover of (a) in the 3D visualization, and (f) is the foldover of (d) in the 3D visualization.}
\label{fig:static_features_and_foldover_features}
\end{figure}

There are two main reasons why the classification results of foldover features are superior to classical static and dynamic features. First, there is a high degree of similarity between different sperms. When two sperms are very similar in shape, size, color and texture, static features cannot effectively distinguish two sperms. However, the foldover features can solve this problem well, because the differences of foldovers between two sperms are very obvious as the example shown in \figurename~\ref{fig:static_features_and_foldover_features}.

According to \figurename~\ref{fig:static_features_and_foldover_features}, we can find that because the sperms in \figurename~\ref{fig:static_features_and_foldover_features} (a) and (d) are very similar in shape, size and color, it is difficult to distinguish them by static features. However, the differences between \figurename~\ref{fig:static_features_and_foldover_features} (a) and (d) on the foldovers are very obvious. 

Second, because sperms are very tiny and there is very little visual information, it is very difficult to distinguish two different sperms. However, due to the foldover features contain not only the original shape and texture information of sperms, but also the movement information of sperms, they can discover more useful visual information. Furthermore, we analyse the X, Y  and Z directions of the sperm foldovers, and expand the sperms movement information, the 3D visualizations of two sperms are shown in \figurename~\ref{fig:static_features_and_foldover_features} (c) and (f).

\begin{figure}[H]
\centering
\includegraphics[scale=0.6]{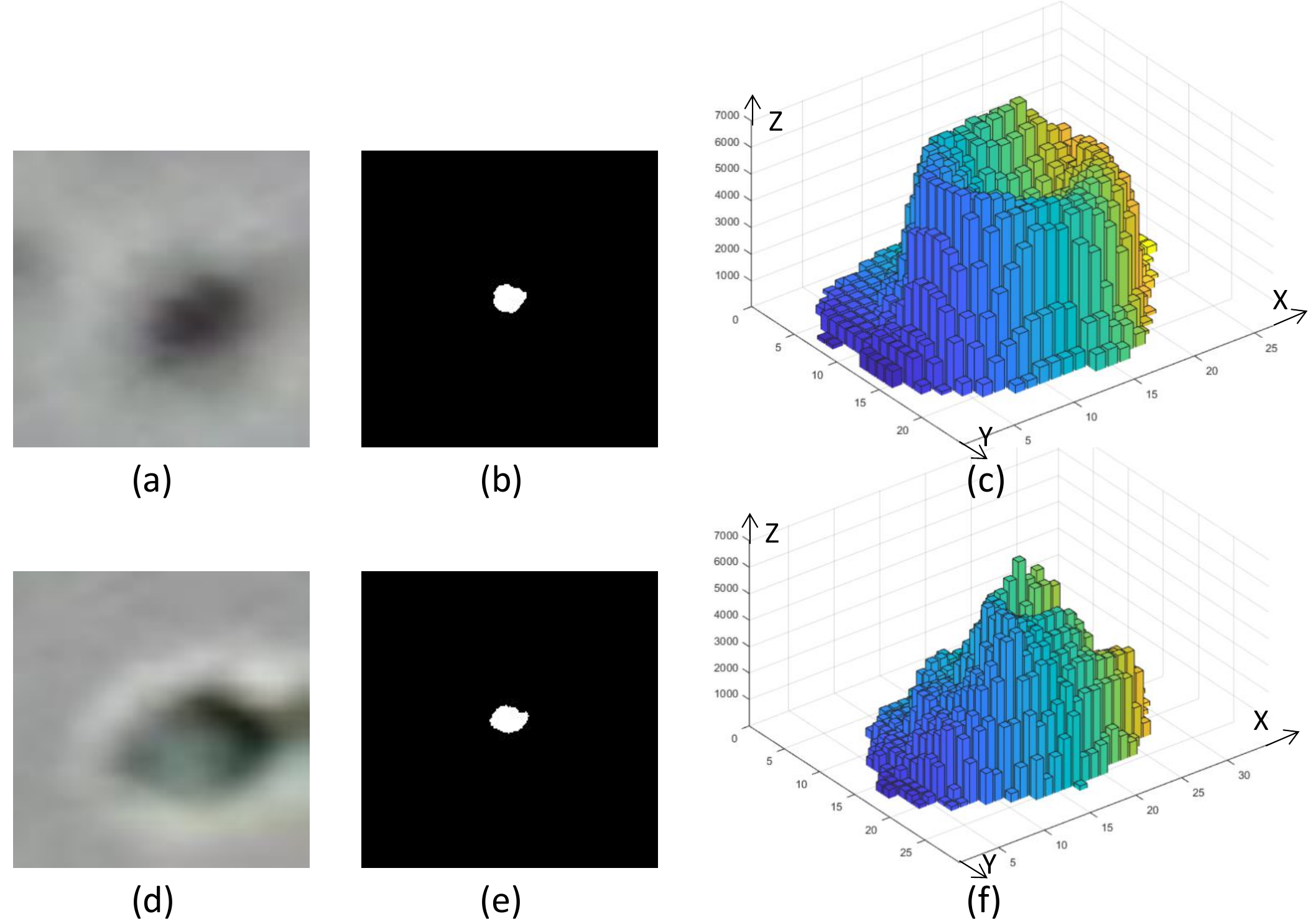}
\caption{Different features of two sperms. (a) and (d) are two different sperms, (b) and (e) are 2D visualization of the two foldovers in the Z direction, and (c) and (f) are 3D visualization of the two foldovers.}
\label{fig:different_feature_two_sperms}
\end{figure}

According to \figurename~\ref{fig:static_features_and_foldover_features} (c) and (f), although \figurename~\ref{fig:static_features_and_foldover_features} (a) and (d) contain little visual information, the information contained in their foldovers is abundant, and the differences between the foldovers are obvious. In addition,  even the static and dynamic features of two sperms are very similar, foldover features contain a lot of visual information to distinguish the sperms as shown in \figurename~\ref{fig:different_feature_two_sperms}.

The two sperms in \figurename~\ref{fig:different_feature_two_sperms} (a) and (d) are very similar in shape, color, size and texture. In \figurename~\ref{fig:different_feature_two_sperms} (b) and (e), there is a high similarity between the two sperms motility states. However, according to the \figurename~\ref{fig:different_feature_two_sperms} (c) and (f), when both static and dynamic features are similar, the information contained in the foldovers is significantly different. It proves that the foldover features are superior in distinguishing tiny objects, similar objects, objects with little visual information and objects with similar visual information.
\section{Conclusion and Future Work}
\label{s:conc}
In this paper, we propose novel foldover features, which are applied to dynamic object behavior description in microscopic videos. Compared with classical static and dynamic features, the foldover features show obvious advantages in distinguishing tiny objects, similar objects, objects with little visual information and objects with similar visual information. In the experiment, we use four different classifiers (ANN, RF, linear-SVM and RBF-SVM) to test the effectiveness of the foldover features, and an overall outstanding classification accuracy is obtained, indicating the effectiveness and potential of the proposed foldover features.

In the future, we plan to increase the amount of data in a single category, allowing the same doctors to expand the data and address the imbalance in our experimental data. Then, although we have tested the foldover features on the semen microscopic videos, we will test it on more highly similar objects to improve the generalization of the foldover features. 
\section{Acknowledgements}
\label{s:ack}
We acknowledge financial support from the “National Natural Science Foundation of China” (No. 61806047) and the “Fundamental Research Funds for the Central Universities” (No. 180719020, N2019003) . We also thank B.E. Frank Kulwa for his great proofreading work.
\bibliographystyle{IEEEtran}
\bibliography{Xialin}

\end{document}